\documentclass[fleqn,10pt]{wlscirep}
\usepackage[utf8]{inputenc}
\usepackage[T1]{fontenc}

\usepackage{adjustbox,lipsum} 

\usepackage{caption}
\usepackage{subcaption}
\usepackage{xurl} 
\usepackage{multirow} 

\usepackage{graphicx}
\usepackage{multirow}
\usepackage{float}
\usepackage{subcaption}
\usepackage{amsmath}
\usepackage{array}
\newcolumntype{C}[1]{>{\centering\arraybackslash}m{#1}}
\usepackage{pifont}

\usepackage{pifont} 

\title{3DAeroRelief: The first 3D Benchmark UAV Dataset for Post-Disaster Assessment}
\author[1\ding{70}]{Nhut Le}
\author[1\ding{70}]{Ehsan Karimi}
\author[1,2,\ding{74}]{Maryam Rahnemoonfar}
\affil[1]{ Department of Computer Science and Engineering, Lehigh University, Bethlehem, Pennsylvania, 18015, USA}
 \affil[2]{Department of Civil and Environmental Engineering, Lehigh University, Bethlehem, Pennsylvania, 18015, USA}
\affil[\ding{74}]{Corresponding author: Maryam Rahnemoonfar (maryam@lehigh.edu)}
\affil[\ding{70}]{These authors contributed equally to this work.}

\begin{abstract}

Timely assessment of structural damage is critical for disaster response and recovery. However, most prior work in natural disaster analysis relies on 2D imagery, which lacks depth, suffers from occlusions, and provides limited spatial context. 3D semantic segmentation offers a richer alternative, but existing 3D benchmarks focus mainly on urban or indoor scenes, with little attention to disaster-affected areas. To address this gap, we present 3DAeroRelief--the first 3D benchmark dataset specifically designed for post-disaster assessment. Collected using low-cost unmanned aerial vehicles (UAVs) over hurricane-damaged regions, the dataset features dense 3D point clouds reconstructed via Structure-from-Motion and Multi-View Stereo techniques. Semantic annotations were produced through manual 2D labeling and projected into 3D space. Unlike existing datasets, 3DAeroRelief captures 3D large-scale outdoor environments with fine-grained structural damage in real-world disaster contexts. UAVs enable affordable, flexible, and safe data collection in hazardous areas, making them particularly well-suited for emergency scenarios. To demonstrate the utility of 3DAeroRelief, we evaluate several state-of-the-art 3D segmentation models on the dataset to highlight both the challenges and opportunities of 3D scene understanding in disaster response. Our dataset serves as a valuable resource for advancing robust 3D vision systems in real-world applications for post-disaster scenarios.

\end{abstract}
\begin{document}

\flushbottom
\maketitle

\thispagestyle{empty}

\section{Background \& Summary}
Timely and accurate post-disaster assessment is essential for effective emergency response, resource allocation, and recovery planning. The current assessment methods primarily rely on 2D satellite and aerial imagery \cite{sarkar21vqaid,SarkarSAMVQA2023,sarkar22gradcam,chowdhury2021attention,Chowdhury21IGARSS,chowdhury21self}, as well as ground-level photography. For instance, datasets like FloodNet \cite{rahnemoonfar2020floodnethighresolutionaerial} and RescueNet \cite{rahnemoonfar2023rescuenet} focus on flooded infrastructure classification and damage severity estimation using 2D imagery. While these datasets provide detailed annotations for 2D analysis, they lack depth information and spatial context, limiting their utility for fine-grained structural assessment in complex disaster scenes.
\\
In contrast, 3D semantic segmentation has emerged as a promising solution for more accurate and comprehensive disaster analysis. Compared to 2D methods, 3D point clouds provide richer spatial context, enabling finer damage assessment and more informed decision-making for emergency response teams. Recent deep learning-based methods \cite{qi2016pointnet, qi2017pointnetplusplus, NEURIPS2018_f5f8590c, Thomas_2019_ICCV, hu2019randla, zhao2021point, wu2022point, Peng_2024_CVPR, wu2024ptv3} have achieved state-of-the-art (SOTA) performance on high-quality indoor \cite{armeni2016s3dis, dai2017scannet, rozenberszki2022language, yeshwanthliu2023scannetpp} and outdoor \cite{behley2019iccv, 9150622, Sun_2020_CVPR, Caesar_2020_CVPR, Liao2022PAMI, gaydon2024fractalultralargescaleaeriallidar} 3D benchmark datasets. For indoor settings, widely used datasets include the Stanford Large-Scale 3D Indoor Spaces (S3DIS) dataset \cite{armeni2016s3dis}, which captures 271 office rooms across six areas; ScanNet \cite{dai2017scannet}, containing over 1,500 RGB-D video sequences; and ScanNet++\cite{Yeshwanth_2023_ICCV}, which incorporates data from laser scanners and DSLR cameras for enhanced geometric and photometric fidelity. These datasets support research in detailed scene understanding but are limited to small-scale and controlled environments. Outdoor datasets, such as SemanticKITTI \cite{behley2019iccv}, Waymo Open\cite{Sun_2020_CVPR}, nuScenes \cite{Caesar_2020_CVPR}, and KITTI-360 \cite{Liao2022PAMI}, offer larger ground-based coverage and are constructed using multi-modal sensors (LiDAR, RADAR, cameras) mounted on autonomous vehicles. However, these ground-based datasets are limited in scale and perspective. They provide only street-level views and often result in sparse point clouds. 
\\
Existing datasets lack disaster-specific annotations and do not represent the unique structural disruptions found in post-disaster scenarios. To address this, we introduce \textbf{3DAeroRelief}, the first 3D benchmark dataset specifically designed for post-disaster assessment using aerial footage captured by unmanned aerial vehicles (UAVs). A comparative summary of existing benchmarks and our proposed dataset is presented in Table~\ref{tab:comp}.  Indoor datasets typically offer high-resolution point clouds from RGB-D scanners but cover only small-scale environments \cite{armeni2016s3dis, dai2017scannet, rozenberszki2022language, Yeshwanth_2023_ICCV}. In contrast, outdoor datasets provide larger coverage but consist of sparse point clouds generated by LiDAR or RADAR, which hinders the ability of models to capture fine structural details \cite{behley2019iccv, Sun_2020_CVPR, Caesar_2020_CVPR, Liao2022PAMI}. Additionally, the reliance on expensive sensing equipment limits deployment in remote or disaster-stricken areas with restricted access to such technology. This highlights a critical gap: the need for large-scale, high-resolution 3D point clouds captured using accessible, low-cost methods. Our dataset fills a major void in the current landscape of 3D semantic segmentation by providing high-resolution data over large-scale disaster areas using low-cost tools. \\
Our main contributions are:

\begin{itemize}
\item \textbf{3D Reconstruction:} We construct high-resolution 3D point clouds from aerial footage captured by UAVs over hurricane-impacted regions in Florida (Hurricane Ian, 2022). The reconstruction leverages Structure-from-Motion (SfM) and Multi-View Stereo (MVS) techniques to generate dense and accurate geometry.

\item \textbf{3D Semantic Annotation:} We develop a pipeline for annotating 3D point clouds by first labeling post-disaster categories in 2D images, then projecting these labels into 3D space. 

\item \textbf{Benchmarking of State-of-the-Art Methods:} We evaluate several state-of-the-art 3D semantic segmentation models on our dataset, analyzing their performance in realistic disaster scenarios and highlighting challenges specific to post-disaster environments.
\end{itemize}

\begin{figure}[!t]
\centering
\begin{subfigure}{0.48\linewidth}
\includegraphics[width=1\linewidth]{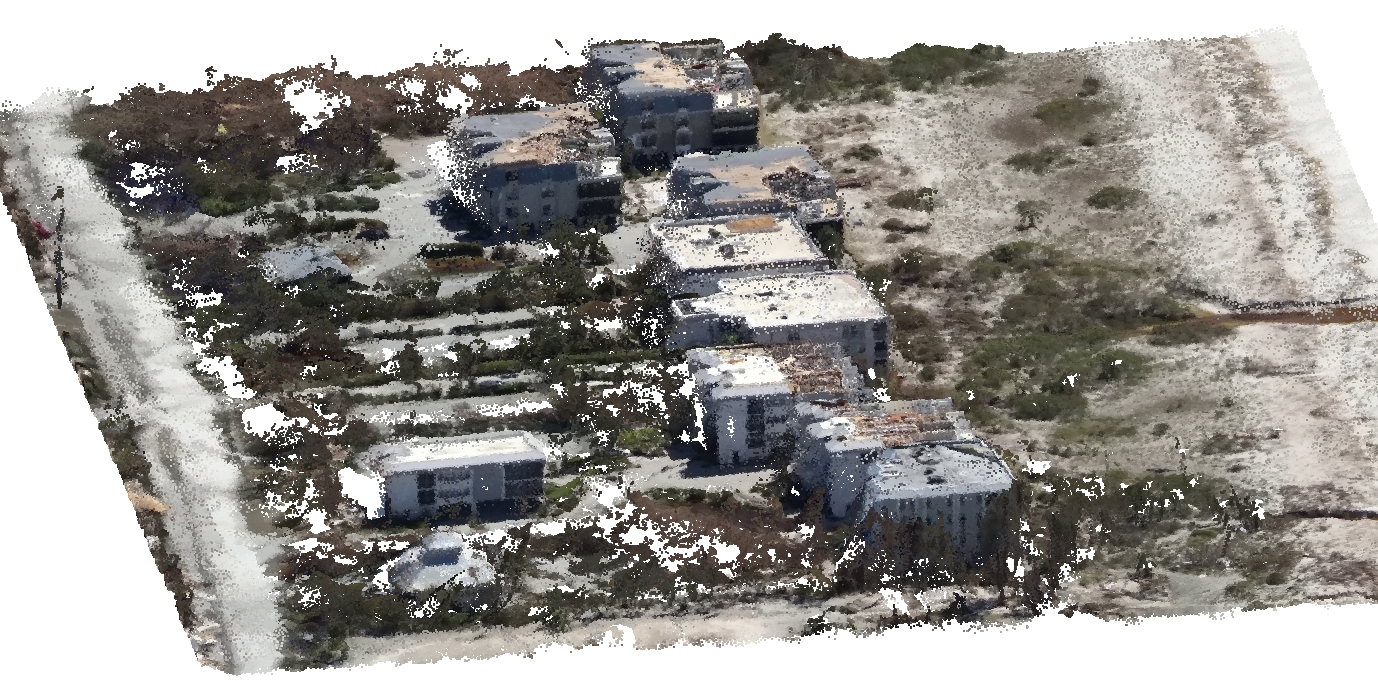}
\end{subfigure}
\hfill
\begin{subfigure}{0.48\linewidth}
\includegraphics[width=1\linewidth]{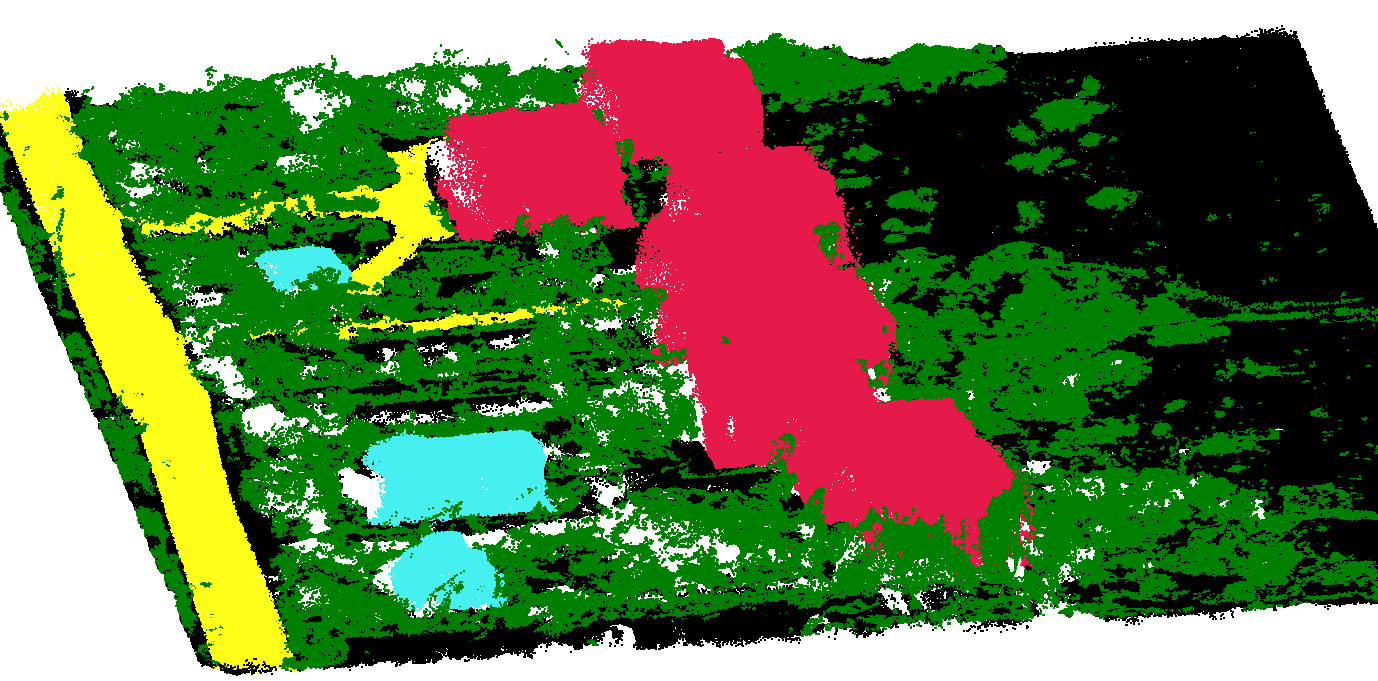}
\end{subfigure}
\hfill
\begin{subfigure}{0.48\linewidth}
\includegraphics[width=1\linewidth]{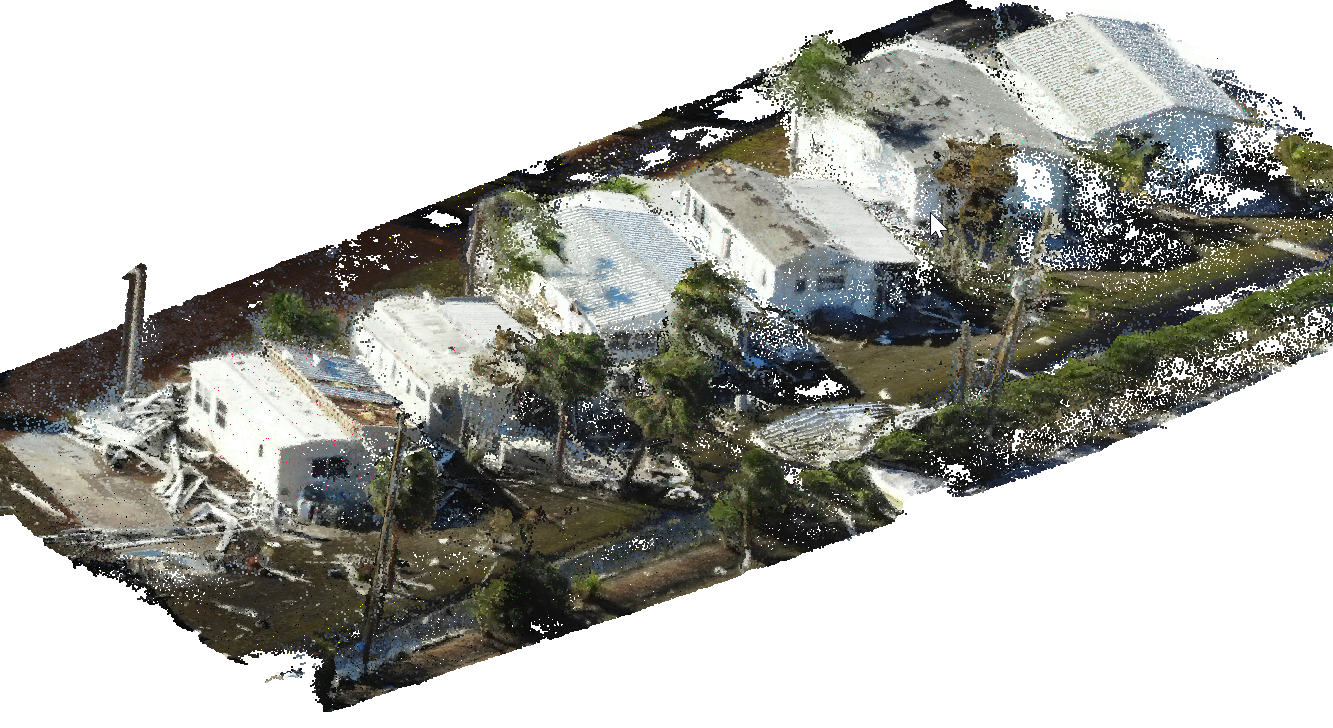}
\end{subfigure}
\hfill
\begin{subfigure}{0.48\linewidth}
\includegraphics[width=1\linewidth]{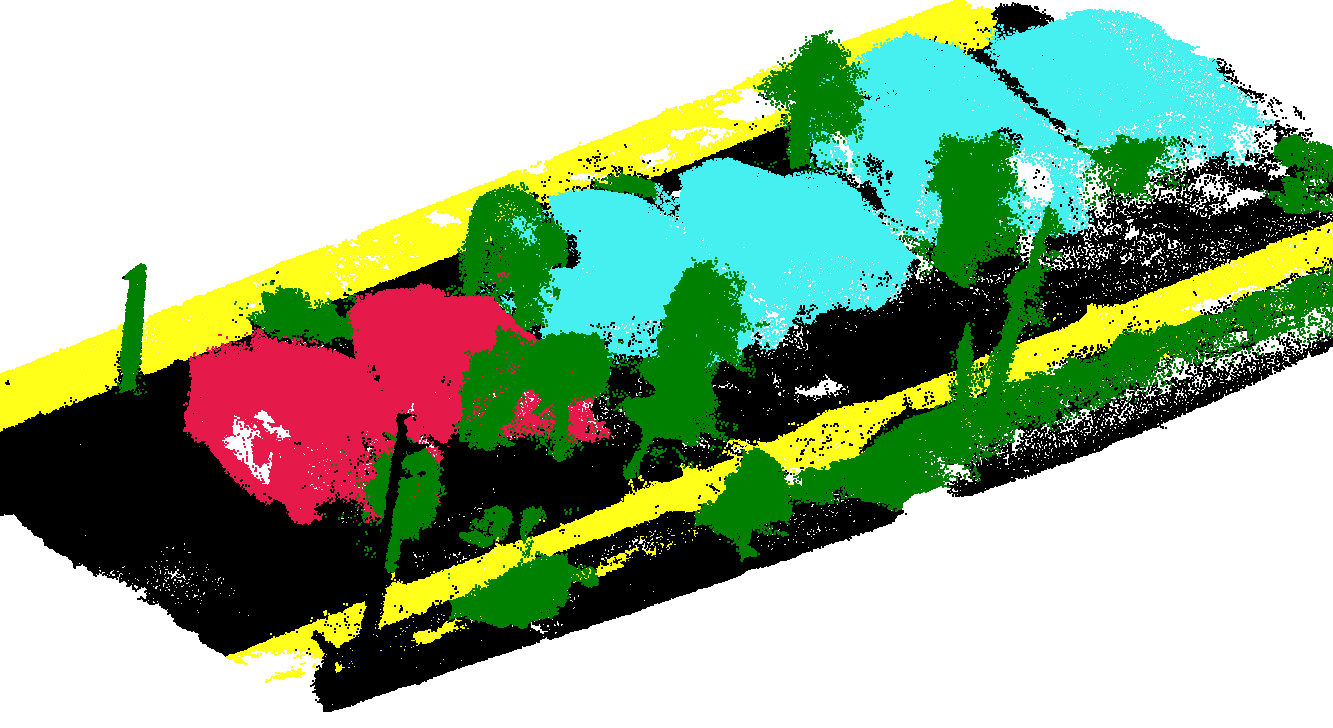}
\end{subfigure}
\hfill
\begin{subfigure}{0.48\linewidth}
\includegraphics[width=1\linewidth]{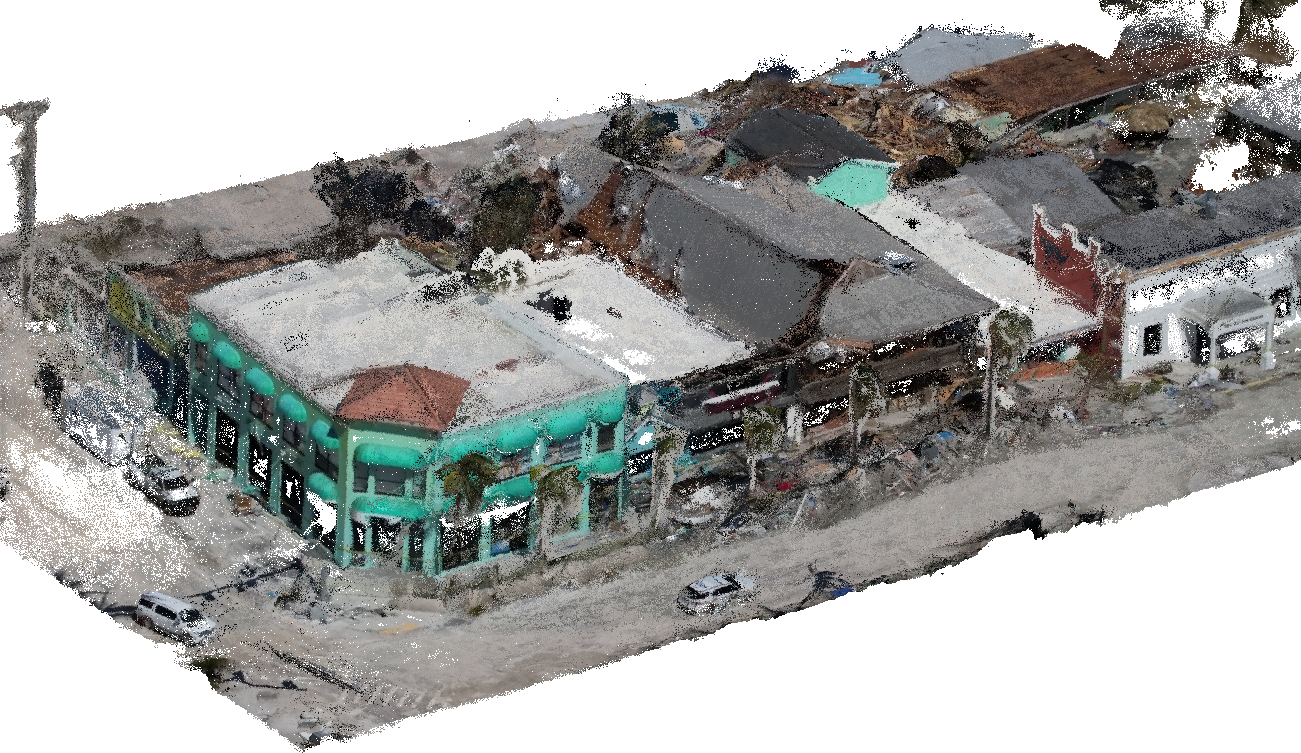}
\end{subfigure}
\hfill
\begin{subfigure}{0.48\linewidth}
\includegraphics[width=1\linewidth]{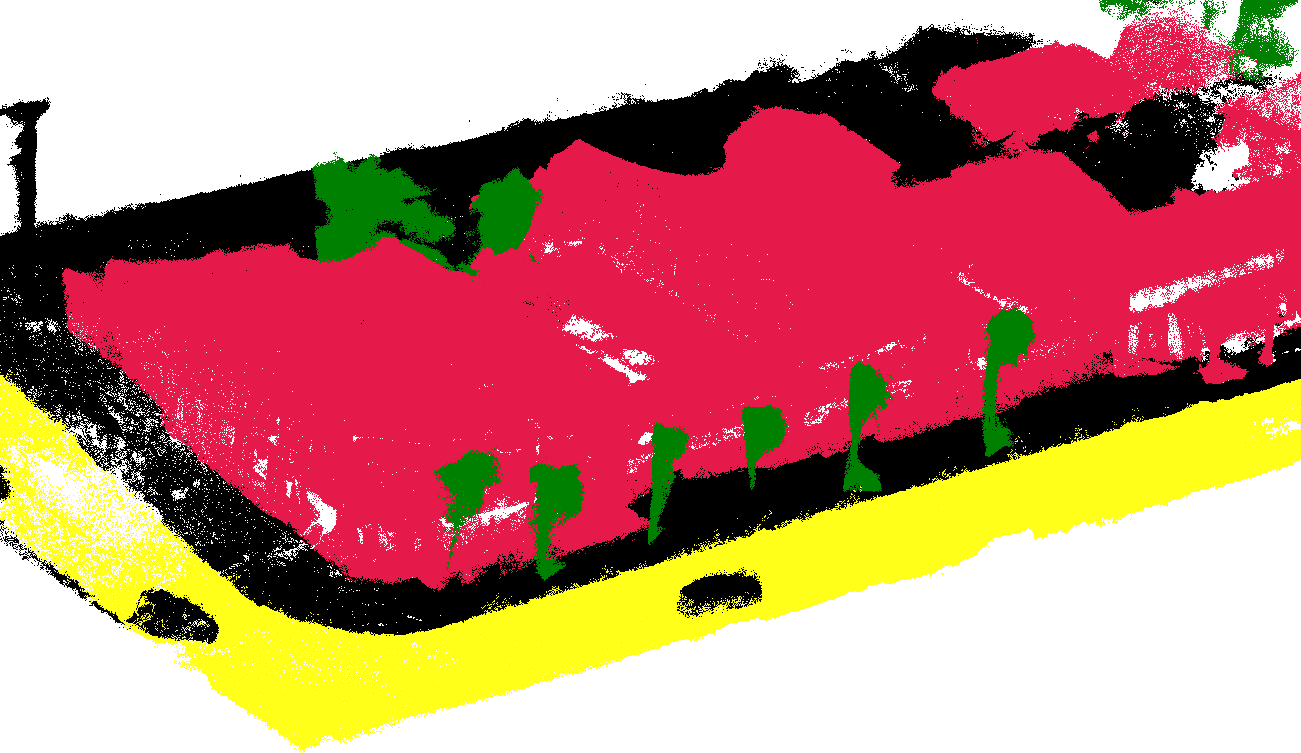}
\end{subfigure}
\caption{Examples of point clouds (left) and their associated semantic labels (right) from our benchmark dataset. \textbf{Cyan:} Building-no-damge, \textbf{Red:} Building-damge, \textbf{Yellow:} Road, \textbf{Green:} Tree, and \textbf{Black:} Background}
\label{fig:dataset}
\end{figure}

\section{Methods}
\begin{figure}[!ht]
\centering
\includegraphics[width=1\linewidth]{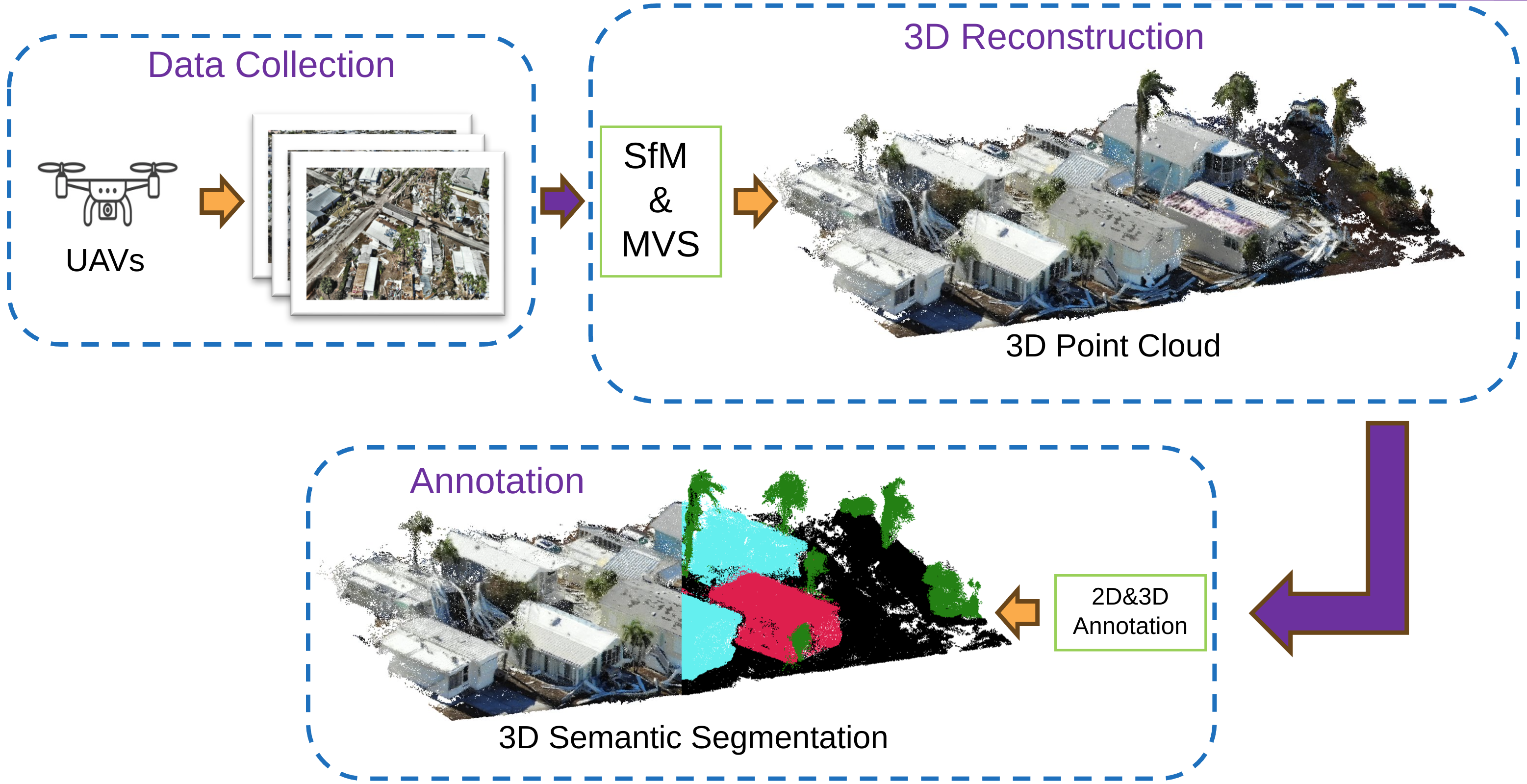}
\caption{\textbf{Dataset Generation Pipeline.} 3DAeroRelief generation consists of three main stages: data collection, 3D reconstruction, and annotation. In the first stage, we collected and pre-processed aerial footage to ensure comprehensive coverage of disaster-affected areas. Frames were then extracted from the footage for use in reconstruction. In the second stage, we applied SfM and MVS techniques to the extracted frames to generate dense 3D point clouds. Finally, in the annotation stage, the reconstructed point clouds were refined, and semantic labels were assigned through manual annotation.}
\label{fig:pipeline}
\end{figure}

\begin{table}[ht]
\centering
\caption{\textbf{Comparison of 3D benchmark datasets.} The datasets differ in data sources, including RGB-D scanners (S3DIS, ScanNet, ScanNet++), LiDAR/RADAR sensors mounted on autonomous vehicles (SemanticKITTI, Waymo, nuScenes, KITTI-360), and UAV-captured RGB imagery (3DAeroRelief). ScanNet++ offers the highest resolution for both indoor and outdoor environments, while 3DAeroRelief (ours) provides the highest-resolution outdoor data specifically tailored for post-disaster assessment.}
\label{tab:comp}

\begin{tabular}{ l C{2cm} C{4cm} C{2cm} C{2cm} C{2cm}}
\hline
\textbf{Dataset} & \textbf{Avg. Points per Scan (millions)} & \textbf{Data Source} & \textbf{Scene Type} & \textbf{Application} & \textbf{Collection Method} \\ \hline \hline
S3DIS & 1.01 & RGB-D scanner & Indoor & Indoor scene understanding & Ground-based\\ 
ScanNet & 1.32 & RGB-D scanner & Indoor & Indoor scene understanding& Ground-based  \\ 
ScanNet200 & 1.32 & RGB-D scanner & Indoor & Indoor scene understanding & Ground-based \\ 
ScanNet++ & \textbf{105.75} & RGB-D scanner & Indoor & Indoor scene understanding & Ground-based \\ \hline
SemanticKITTI & 0.0987 & LiDAR from AVs & Outdoor & Autonomous driving & Ground-based\\ 
Waymo & -- & LiDAR from AVs & Outdoor & Autonomous driving& Ground-based \\ 
nuScenes & 0.041 & LiDAR + RADAR from AVs & Outdoor & Autonomous driving& Ground-based \\ 
KITTI-360 & -- & LiDAR from AVs & Outdoor & Autonomous driving& Ground-based \\ 
\textbf{3DAeroRelief (Ours)} & \textbf{0.520} & \textbf{UAV RGB images} (via SfM + MVS reconstruction) & Outdoor & \textbf{Post-disaster assessment} & \textbf{Aerial}\\ \hline
\end{tabular}
\end{table}
To the best of our knowledge, no existing 3D benchmark dataset has been specifically developed for post-disaster environments. To address this gap, we introduce \textbf{3DAeroRelief}: a 3D dataset specifically designed for the unique challenges of disaster-affected areas. Figure \ref{fig:dataset} illustrates the 3D point clouds and their associated semantic labels in our dataset. Inspired by prior 2D efforts such as FloodNet \cite{rahnemoonfar2020floodnethighresolutionaerial} and RescueNet \cite{rahnemoonfar2023rescuenet}, aerial footage of regions in Florida impacted by Hurricane Ian (2022) is collected by UAVs. We then applied SfM \cite{schoenberger2016sfm} and MVS \cite{schoenberger2016mvs} techniques to reconstruct dense 3D point clouds of the affected areas. For semantic ground-truth labels, 2D images were manually annotated and then projected into 3D space, enabling the assignment of semantic labels to corresponding points in the reconstructed scenes. Final semantic labels are refined using interactive 3D editing tools \cite{cloudcompare} to ensure high-quality point-level annotations. Figure \ref{fig:pipeline} illustrates the pipeline of our 3D benchmark dataset generation.
\subsection{Data Collection and Pre-processing}
\subsubsection{Data Collection} To ensure comprehensive coverage of the affected regions, high-resolution aerial footage was captured using small UAVs. The data was collected by the Center for Robot-Assisted Search and Rescue (CRASAR) on behalf of the Florida State Emergency Response Team during the disaster response phase. This real-time collection by emergency responders ensures that the dataset accurately reflects real-world post-disaster conditions.

\begin{figure}[!t]
\centering
\begin{subfigure}{0.48\linewidth}
\includegraphics[width=1\linewidth]{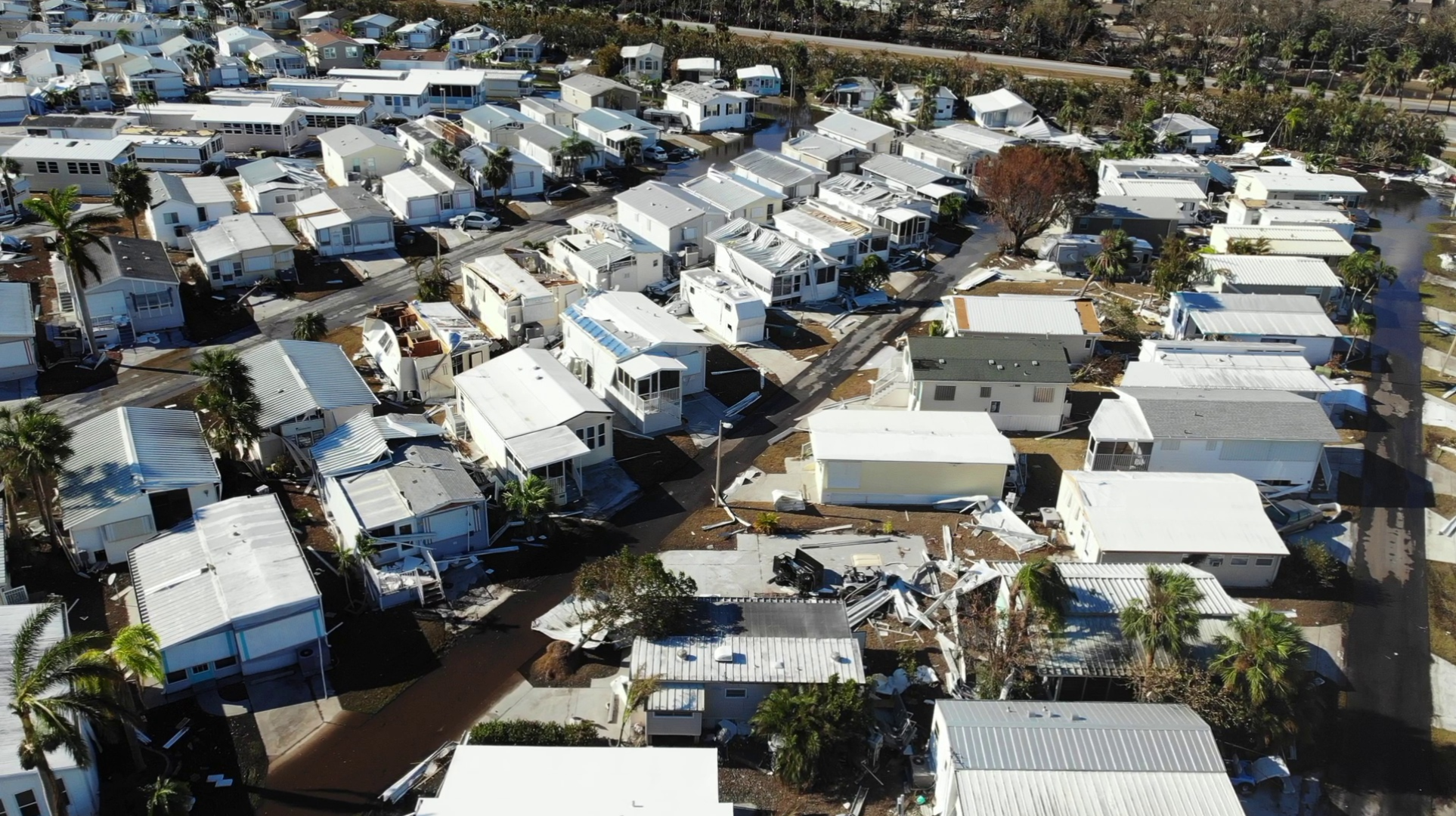}
\end{subfigure}
\hfill
\begin{subfigure}{0.48\linewidth}
\includegraphics[width=1\linewidth]{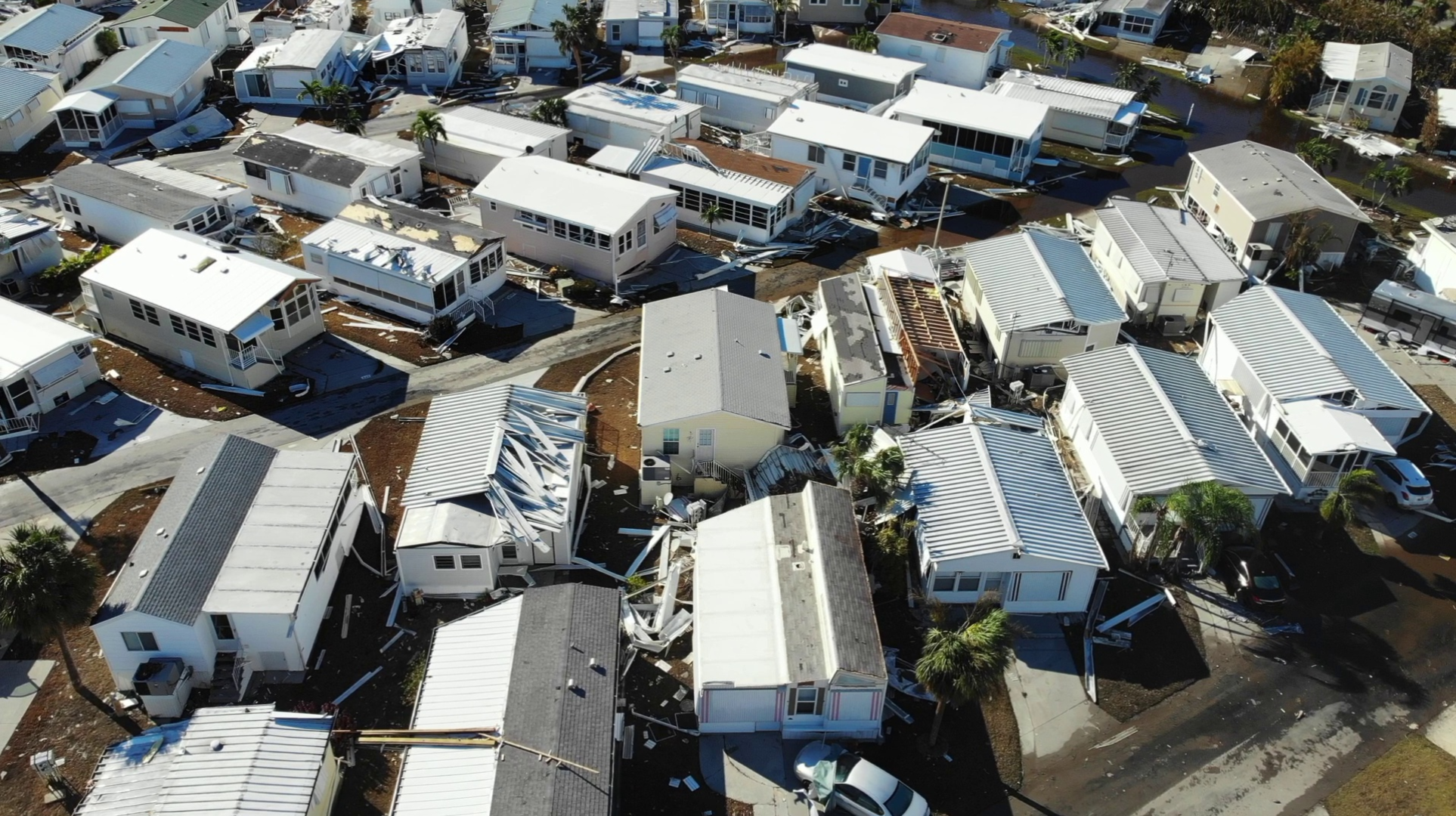}
\end{subfigure}
\hfill
\begin{subfigure}{0.48\linewidth}
\includegraphics[width=1\linewidth]{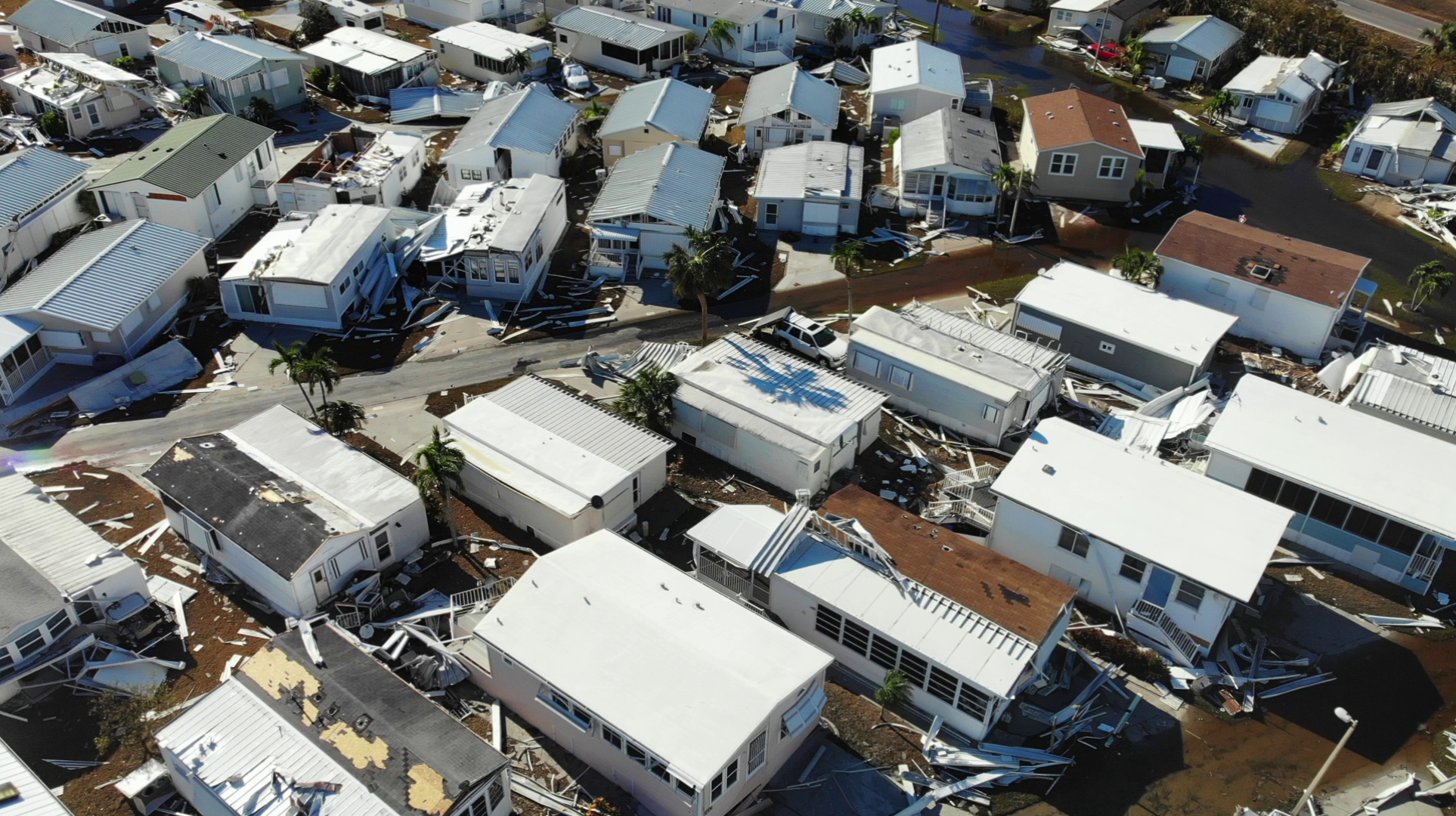}
\end{subfigure}
\hfill
\begin{subfigure}{0.48\linewidth}
\includegraphics[width=1\linewidth]{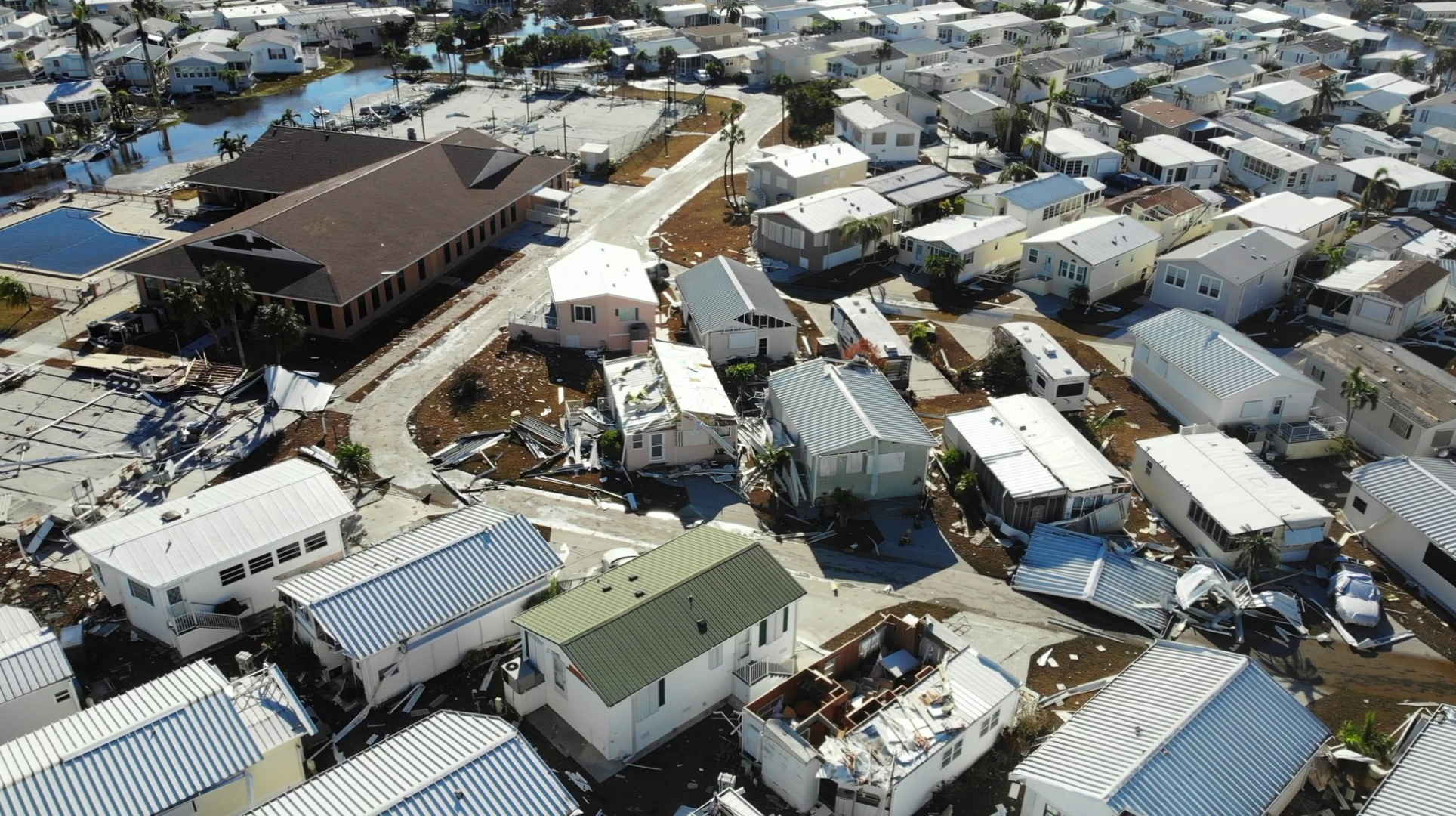}
\end{subfigure}
\caption{\textbf{Examples of Extracted Frames from Aerial Footage} Frames were extracted from the pre-processed aerial footage, with each frame having a resolution of $1920 \times 1080$ pixels.}
\label{fig:frame}
\end{figure}

\subsubsection{Pre-processing} The aerial video footage was carefully reviewed to remove irrelevant or low-quality content, reducing noise and computational overhead in the 3D reconstruction stage. The cleaned footage was then segmented into individual frames, each with a resolution of $1920 \times 1080$ pixels, which served as input for the reconstruction pipeline. Figure~\ref{fig:frame} illustrates examples of the extracted frames.


\subsection{3D Reconstruction}
We reconstruct 3D point clouds from aerial RGB images using a two-stage pipeline comprising Structure-from-Motion (SfM) \cite{schoenberger2016sfm} and Multi-View Stereo (MVS) \cite{schoenberger2016mvs}. The input consists of extracted frames from UAV footage covering disaster-affected regions. SfM  \cite{schoenberger2016sfm} builds a sparse 3D structure and estimates camera parameters, while MVS  \cite{schoenberger2016mvs} densifies the model into a high-resolution 3D point cloud.

\paragraph{Structure-from-Motion (SfM)}
SfM \cite{schoenberger2016sfm} recovers both the 3D geometry of a scene and the intrinsic and extrinsic camera parameters from unordered image collections. The pipeline proceeds in three key stages:

\begin{enumerate}
\item \textbf{Feature Matching and Geometric Filtering:} Keypoints are matched between image pairs to establish 2D-2D correspondences. To eliminate incorrect matches, geometric constraints are applied:
\begin{itemize}
    \item \textit{Homography} is used to model the relationship between two views of a planar surface or a scene with limited depth variation. It ensures that points lying on the same plane in one image correspond to the same plane in another.
    \item \textit{Epipolar Geometry} describes the geometric relationship between two views of the same scene. It defines the epipolar constraint: given a point in one image, its corresponding point in the second image must lie along a specific line--the epipolar line--determined by the relative pose between the two cameras. Matches violating this constraint are considered outliers and removed using RANSAC.
\end{itemize}

\item \textbf{Incremental 3D Reconstruction:} The process starts from a well-matched image pair to initialize the reconstruction. Additional images are incrementally registered using the \textit{Perspective-n-Point (PnP)} algorithm, which estimates the camera pose of a new image from known 3D points and their corresponding 2D projections. As each new view is added, new 3D points are triangulated by computing the intersection of corresponding rays back-projected from image pixels. The entire reconstruction is iteratively refined using bundle adjustment to jointly optimize camera parameters and 3D point positions, minimizing reprojection error across all observations.
\end{enumerate}

\paragraph{Multi-View Stereo (MVS)}
MVS \cite{schoenberger2016mvs} densifies the sparse point cloud generated by SfM \cite{schoenberger2016sfm} by estimating depth information for each pixel across multiple views. This stage reconstructs detailed 3D surfaces as follows:

\begin{enumerate}
\item \textbf{Depth Estimation:} For each reference image, dense depth maps are generated by leveraging \textit{photometric consistency}, which assumes that the same 3D point will project to similar pixel intensities across overlapping views. Pixel-level matching is performed across image pairs to identify consistent depth hypotheses.

\item \textbf{Search Space Optimization:} The space of depth candidates is constrained and refined using techniques such as plane sweeping, which evaluates the photometric similarity of projected pixels across planes at different depths. Multi-scale filtering is also applied to reduce noise and improve stability across textureless regions.

\item \textbf{Dense Point Cloud Generation:} Final depth maps from multiple views are fused into a single dense 3D point cloud. Confidence weighting and consistency checks ensure that only reliable depth estimates contribute to the final geometry.
\end{enumerate}

\subsection{Post-Processing}
Although the above process can generate dense 3D point clouds from image sequences, it often introduces outliers due to mismatches between keypoints across frames. To address this, a post-processing step is required to refine and clean the reconstruction. This includes aligning the point cloud to an absolute scale and isolating specific regions of interest for further usage.
\subsubsection{Absolute Re-scaling}
\begin{figure}[!ht]
\centering
\includegraphics[width=0.7\linewidth]{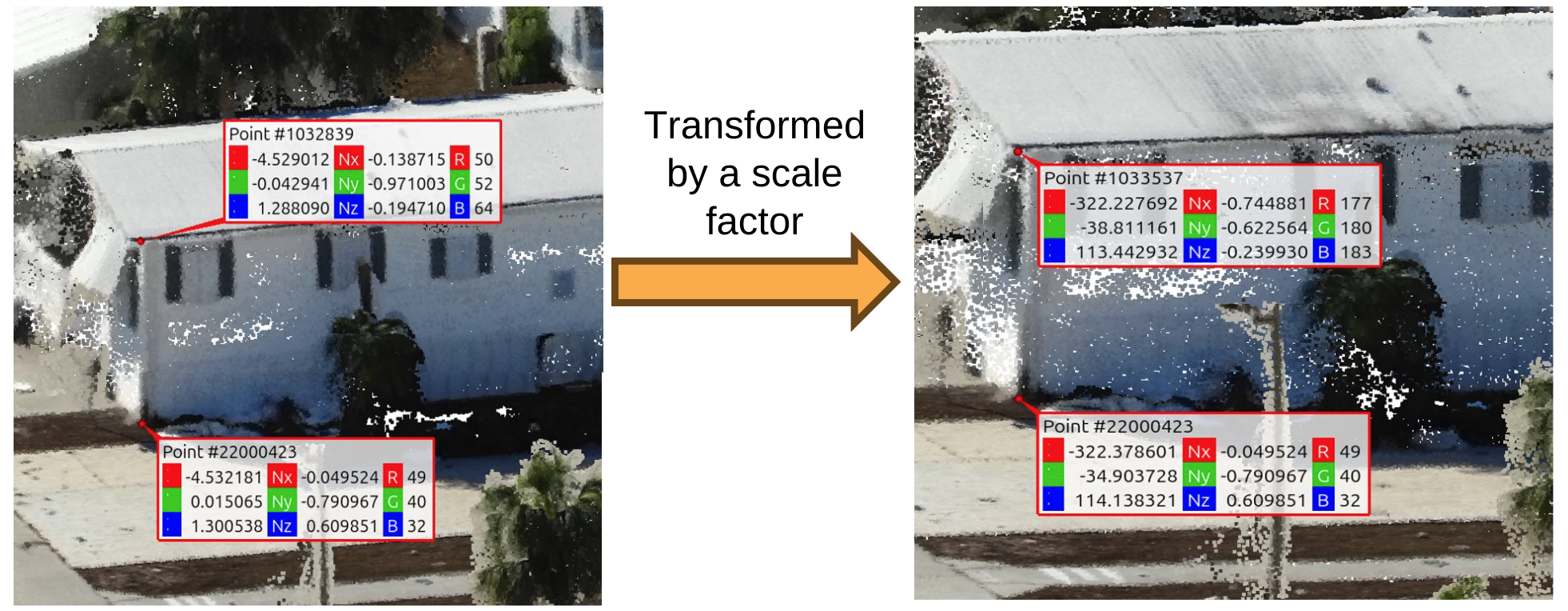}
\caption{\textbf{The Rescaling Process} First, two points representing the height of a house are manually selected in the reconstructed point cloud (\textbf{left image}). The scale factor $s$ is then computed based on the known real-world height. Finally, this factor is applied uniformly to the entire point cloud to obtain an approximately scaled 3D reconstruction in metric units (\textbf{right image}).}
\label{fig:rescale}
\end{figure}

Most 3D semantic segmentation networks require input data to be aligned with the real-world scale. However, SfM \cite{schoenberger2016sfm} produces reconstructions with arbitrary scale, as it relies solely on relative image-based feature matching. To convert the reconstruction into metric units, we apply a rescaling procedure using a known reference, the average height of a single-story house ($h_{\text{real}} \approx 3$ meters).
\paragraph{Rescaling Methodology}
Figure~\ref{fig:rescale} illustrates the rescaling process. We first manually identify two 3D points corresponding to the vertical extent of a single-story house in the reconstructed point cloud. Let $h_{\text{recon}}$ be the measured height between these two points in the original (relative-scale) reconstruction. The scale factor is then computed as $s = h_{\text{real}} / h_{\text{recon}}$. This factor is applied uniformly to scale the entire point cloud about its centroid, aligning it with real-world dimensions. 

\subsubsection{Cleaning and Isolating Regions of Interest}
\begin{figure}[!ht]
\centering
\includegraphics[width=0.7\linewidth]{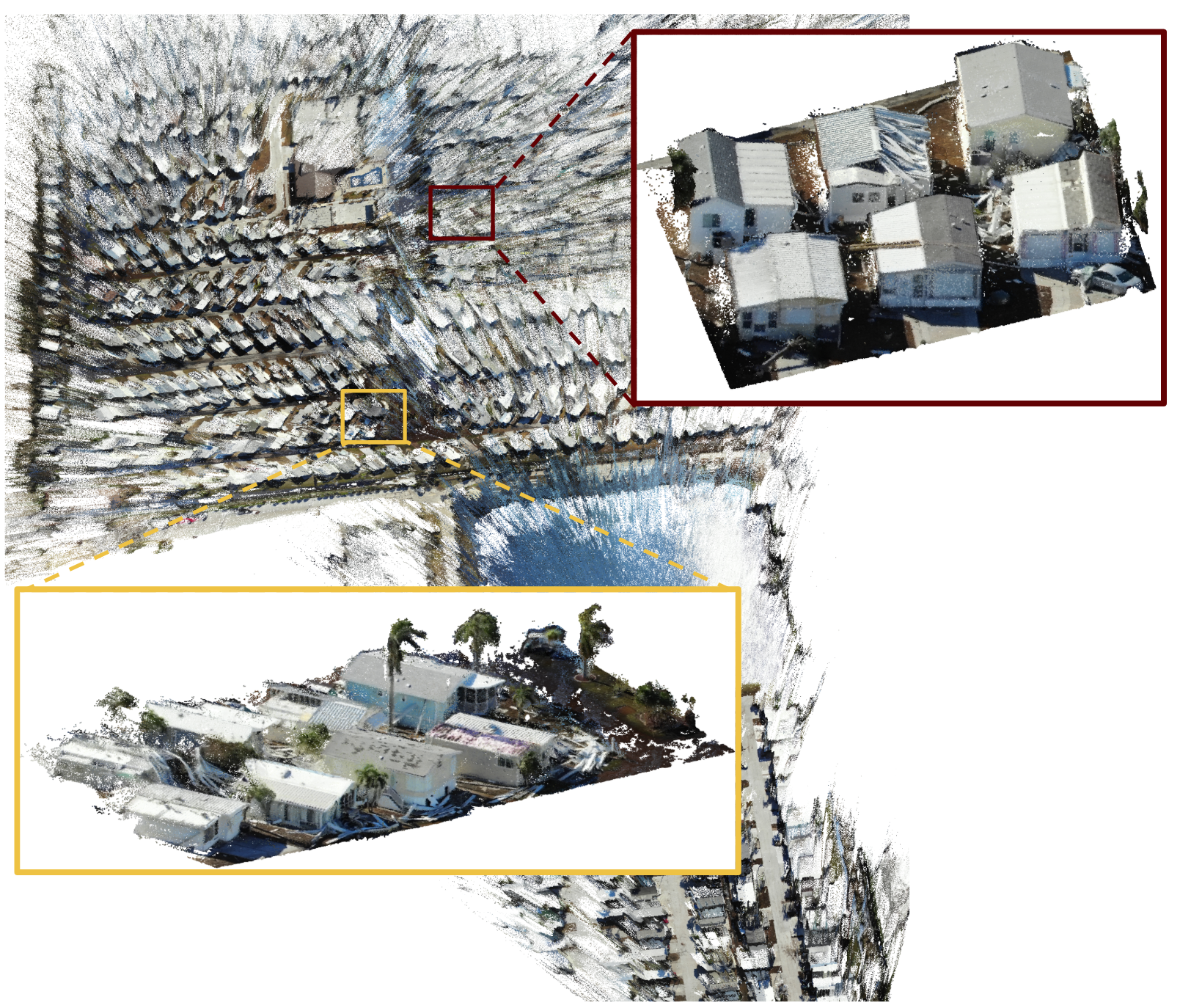}
\caption{\textbf{Example of the Cleaning and Isolating Regions of Interest process} Two point cloud segments are extracted from the large, noisy reconstruction. Each is cleaned and refined to provide a clearer and more focused representation of the disaster-affected scene.}
\label{fig:selectpcd}
\end{figure}

To support effective training of 3D semantic segmentation models, it is important to construct a dataset with a balanced representation of both damaged and undamaged structures. The raw reconstructed point clouds often contain large areas dominated by undamaged buildings, which could lead to class imbalance and hinder the model's ability to learn fine-grained features of post-disaster damage. Therefore, we selectively isolate regions containing a representative mix of damaged and undamaged structures. This targeted selection ensures a more balanced dataset, improves training effectiveness, and focuses learning on the most relevant features for post-disaster assessment. In addition, this process reduces annotation overhead and computational demands. Figure~\ref{fig:selectpcd} shows examples of two cleaned point cloud segments extracted from the larger reconstruction.
\subsection{Annotation}
Accurate ground-truth data is essential for training reliable deep learning models in 3D semantic segmentation. To ensure high-quality annotations, we adopted a three-step process: (1) manual annotation of 2D images, (2) projection of these labels into 3D space to generate semantic labels for point clouds, and refinement of the 3D annotations using dedicated 3D editing software, such as CloudCompare \cite{cloudcompare}.
\subsubsection{2D Annotation}
To minimize annotation effort while maintaining high labeling quality, we manually annotated every 10th of order frame in the dataset. Each object was labeled with a semantic class, following a scheme largely consistent with the methodology used in RescueNet \cite{rahnemoonfar2023rescuenet}. The complete list of semantic categories is provided in Table~\ref{tab:instance}.

    \begin{table}[ht!]
    \caption{Instance types and classes in our 3D benchmark dataset} 
\label{tab:instance}
\centering
\begin{tabular}{ c l l }
\hline
\textbf{Instance Type}    & \multicolumn{1}{c }{\textbf{Classes}} & \multicolumn{1}{c }{\textbf{Definition}}          \\ \hline\hline
\multirow{2}{*}{Building} & Building-no-damage                    & Buildings with minimal or no visible structural damage  \\ \cline{2-3} 
                          & Building-damage                 & Buildings exhibiting structural damages                                 \\ \hline
Road     & Road                            & Roads that are at any condition 
                                  \\ \hline
Tree                      & Tree                                  & Individual trees or clusters of trees                          \\ \hline
Background                & Background                            & Remaining unclassified objects and terrain                             \\ \hline
\end{tabular}
\end{table}
\subsubsection{3D Annotation and Refinement}
Following the 2D annotation step, labels were projected into 3D space to generate ground-truth segmentation for the reconstructed point clouds. Each 3D point was reprojected into all annotated 2D frames using known camera intrinsics and extrinsics obtained from the SfM \cite{schoenberger2016sfm} reconstruction. Specifically, given a 3D point $\mathbf{X} = [X, Y, Z, 1]^T$ in homogeneous coordinates and a camera projection matrix $P = K[R|t]$, where $K$ is the intrinsic matrix and $[R|t]$ represents the extrinsic transformation, the corresponding 2D image coordinates $\mathbf{x} = [u, v]^T$ are obtained via:
$$\mathbf{x} \sim P\cdot \mathbf{X} = K \cdot (R\mathbf{X} + \mathbf{t})$$
For each 3D point, we collect all valid 2D projections that fall within the bounds of an annotated image. Using these projections, we assign the most frequent semantic label (via majority voting) from the overlapping 2D annotations. This multi-view strategy improves labeling robustness by aggregating information from different viewpoints. Thus, it mitigates errors due to occlusion or annotation noise. Finally, manual refinements were performed using the 3D tool \cite{cloudcompare} to correct label inconsistencies and ensure accurate point-level annotations. 

\subsection{Dataset Splits}
\textbf{3DAeroRelief} consists of sixty-four 3D point clouds collected from eight distinct areas, with each point cloud containing at least one building. The primary point clouds include at least one damaged building to ensure balanced representation and a strong emphasis on post-disaster scenarios. For evaluation, we employ a cross-area strategy: all baseline models are trained from scratch using fifty-six point clouds from Areas 1, 3, 4, 5, 6, 7, and 8, and evaluated on eight point clouds from Area 2. This split is designed to assess the generalization capability of 3D segmentation methods on previously unseen, disaster-affected regions.
\section{Data Records}
The dataset is organized as follows:
\begin{itemize}
    \item \textbf{Area\_<n> directories:} Each directory corresponds to a specific area and contains point clouds named \textit{pp<i>.ply}, where $i$ is the index. The corresponding semantic segmentation files are named \textit{segmentpp<i>.ply}.
    \item \textbf{labels.txt:} A text file listing the semantic class labels along with their associated RGB color codes.
\end{itemize}
\section{Technical Validation}
\subsection{Evaluation Metrics}
We evaluate 3D semantic segmentation performance on two standard metrics: \textbf{Mean Class Accuracy (mAcc)} and \textbf{Mean Intersection over Union (mIoU)}. Let $N$ be the total number of classes, $\text{TP}_i$ the number of true positives, $\text{FP}_i$ the number of false positives, and $\text{FN}_i$ the number of false negatives for class $i$.
\begin{itemize}
    \item \textbf{Mean Class Accuracy (mAcc):}
    \[
    \text{mAcc} = \frac{1}{N} \sum_{i=1}^{N} \frac{\text{TP}_i}{\text{TP}_i + \text{FN}_i}
    \]
    This metric measures the average classification accuracy across all classes.
    
    \item \textbf{Mean Intersection over Union (mIoU):}
    \[
    \text{mIoU} = \frac{1}{N} \sum_{i=1}^{N} \frac{\text{TP}_i}{\text{TP}_i + \text{FP}_i + \text{FN}_i}
    \]
    This metric quantifies the average overlap between predicted and ground-truth segments for each class.
\end{itemize}
\subsection{Baselines}
\begin{table*}[!ht]
\caption{ \textbf{mIoU} on our 3D dataset for SOTA 3D semantic segmentation methods}
\centering
\begin{tabular}{c|c|ccccc}
\textbf{Methods} & \textbf{Overall} & \textbf{Building-Damage} & \textbf{Building-no-Damage} & \textbf{Road} & \textbf{Tree} & \textbf{Background} \\ \hline\hline

PTv1             &  0.3084   &  0.5112   & 0.1543      & 0.3002 &0.1909  &   0.3854       \\
PTv2      & 0.2559  & 0.0020  &  \textbf{0.2531}   &    \textbf{0.4864}     &  0.0367   &   0.5014     \\
PTv3      &  \textbf{0.4584}     & \textbf{0.7614}        & 0.3108                                                                 & 0.0467                                                             & \textbf{0.4001}        & \textbf{0.7730 }                   \\
FPT &   0.11  &   0.14  &   0.03  & 0.00 &   0.00  & 0.41   \\
OA-CNNs          & 0.2604&  0.5822 &  0.1962   &  0.00   & 0.2959 &      0.2277   
\end{tabular}
\label{tab:eval-miou}
\end{table*}
\begin{table*}[!ht]
\caption{ \textbf{mAcc} on our 3D dataset for SOTA 3D semantic segmentation methods}
\centering
\begin{tabular}{c|c|ccccc}
\textbf{Methods} & \textbf{Overall} & \textbf{Building-Damage} & \textbf{Building-no-Damage} & \textbf{Road} & \textbf{Tree} & \textbf{Background} \\ \hline\hline

          
PTv1 &         0.4606      & 0.7650  & 0.3010  &    0.3002           &   0.5071        & 0.4297 \\
          
PTv2 &   0.3941    &    0.0020           & \textbf{0.4826}  &    \textbf{0.5152 }         &  0.3628         &   0.6077        \\
          
PTv3 &              \textbf{0.5508}& 0.8949&                                                                  0.4797&               0.0910&               0.4149&                    \textbf{0.8734}\\
          
FPT &    0.22  &   0.14  &  0.42  & 0.00 &   0.00  & 0.53    \\
                      
                      
                      
OA-CNNs &  0.4396 &  \textbf{0.9276}  &  0.3030    &   0.00      &   \textbf{0.7248}   &        0.2427          \\
            
\end{tabular}
\label{tab:eval-macc}
\end{table*}
We evaluate our 3D dataset using the following SOTA 3D semantic segmentation methods: 
Point Transformer (PTv1) \cite{zhao2021point},  Point Transformer v2 (PTv2) \cite{wu2022point}, Point Transformer v3 (PTv3) \cite{wu2024ptv3}, Fast Point Transformer (FPT) \cite{park2022fast}, 
 and Omni-Adaptive sparse CNNs (OA-CNNs) \cite{Peng_2024_CVPR}.
Inspired by the success of self-attention mechanisms in natural language processing and 2D vision, PTv1 \cite{zhao2021point} introduces a self-attention-based architecture designed for unordered 3D point sets. PTv2 \cite{wu2022point} enhances PTv1 \cite{zhao2021point} with Grouped Vector Attention, a refined position encoding strategy, and a novel partition-based pooling mechanism. PTv3 \cite{wu2024ptv3} further builds on PTv2 \cite{wu2022point} by introducing point cloud serialization - transforming unstructured point clouds into structured sequences via uniform grids and space-filling curves (e.g., Z-order, Hilbert). This enables transformer layers to operate without the need for costly neighbor searches, such as k-nearest neighbors. PTv3 \cite{wu2024ptv3} also introduces xCPE, a lightweight sparse convolutional module that efficiently captures spatial relations without complex relative positional encodings. FPT \cite{park2022fast} introduces an attention-based framework for 3D semantic segmentation that leverages centroid-aware voxelization and devoxelization to incorporate a learnable centroid-to-point positional encoding. By voxelizing the input point cloud, it can efficiently process large-scale point clouds with only a minor decrease in segmentation accuracy. 
OA-CNNs \cite{Peng_2024_CVPR} tackles the limitations of traditional sparse CNNs in handling variable local geometries. It introduces two core components: Adaptive Receptive Fields, which adjust field sizes based on geometric context, and Adaptive Relation Mapping, which dynamically models local feature relationships without relying on self-attention. These innovations enable OA-CNNs to capture fine details in complex regions while maintaining global context, achieving competitive performance with transformer-based models like PTv2 \cite{wu2022point}.
\subsection{Analysis}
{
\setlength{\tabcolsep}{1pt}
\renewcommand{\arraystretch}{0.1}
\begin{table}[!ht]
\caption{\textbf{Qualitative Results.} The outputs of 
PTv1 \cite{zhao2021point}, FPT \cite{park2022fast}, PTv2 \cite{wu2022point}, 
PTv3 \cite{wu2024ptv3}, and OA-CNNs \cite{Peng_2024_CVPR} on the test set. Building-no-Damage (Cyan), Building-Damage (Red), Tree (Green), Road (Yellow), and Background (Black)}
\centering
\begin{tabular}{cccc}
 & \includegraphics[width=0.23\linewidth]{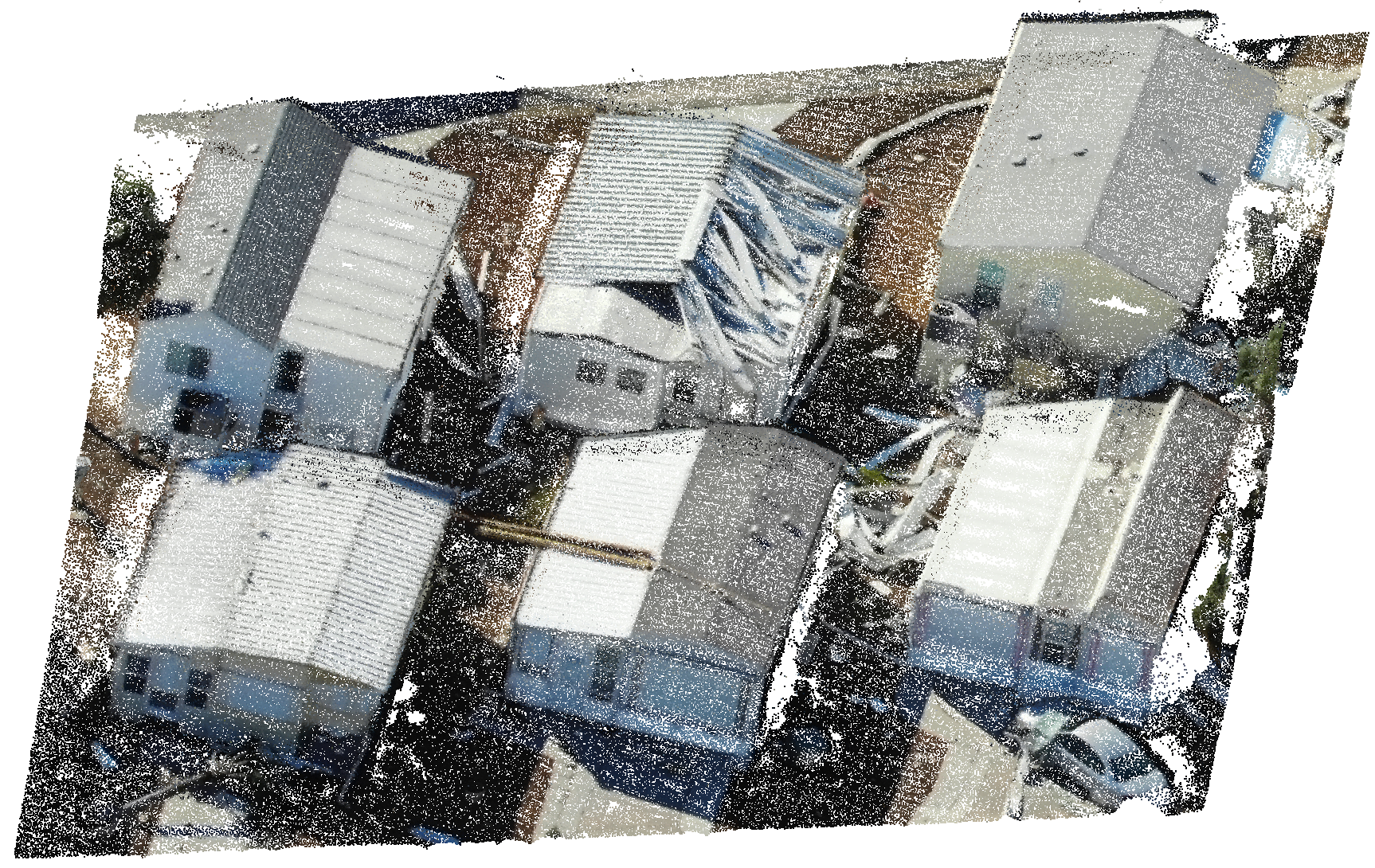}  & \includegraphics[width=0.3\linewidth]{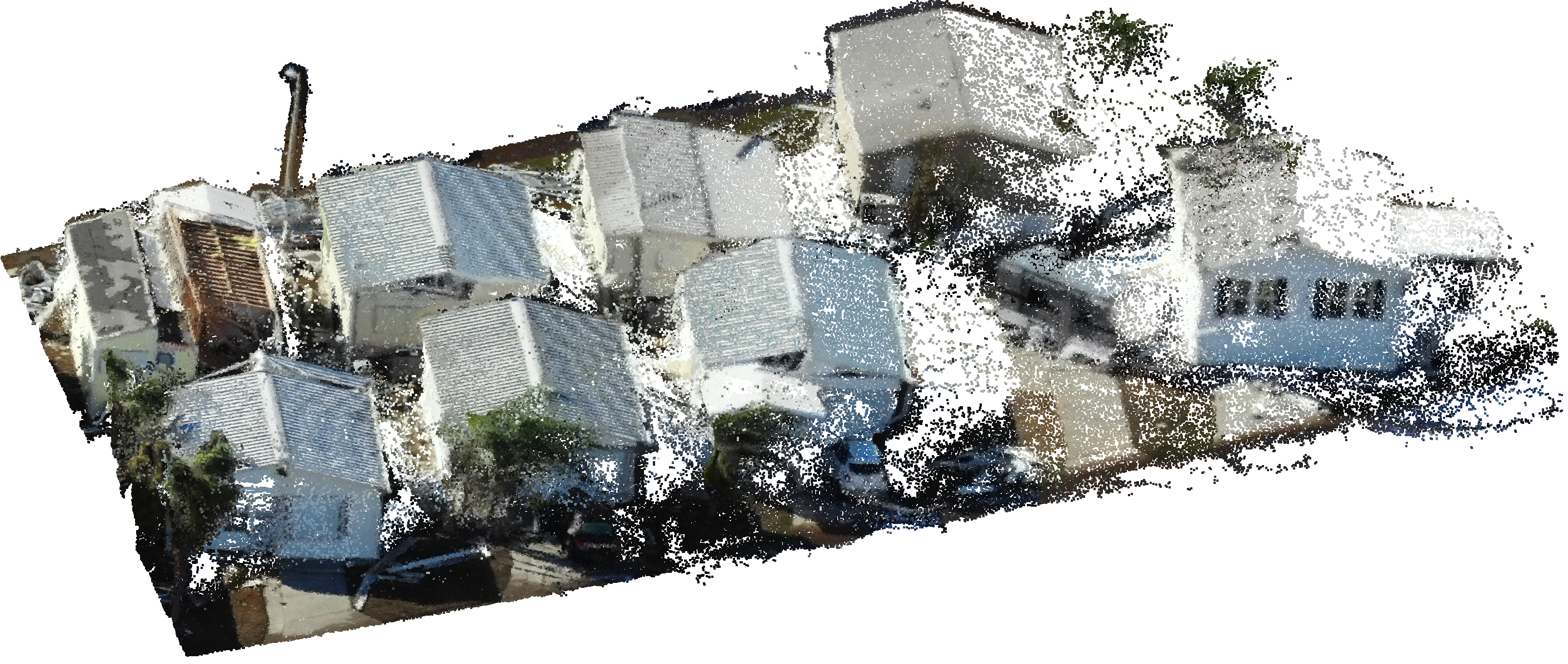} & \includegraphics[width=0.23\linewidth]{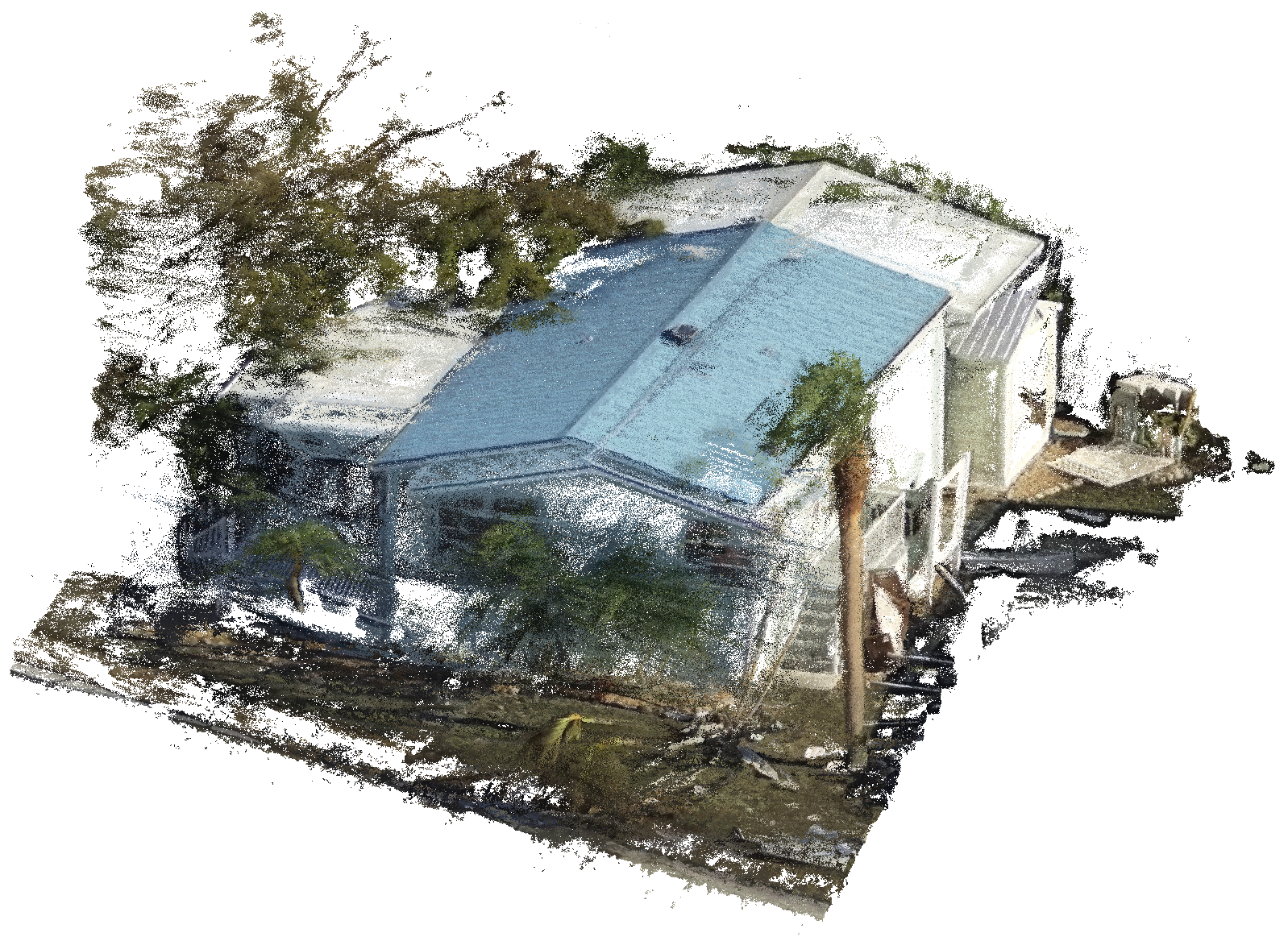} \\ 
Ground Truth & \includegraphics[width=0.23\linewidth]{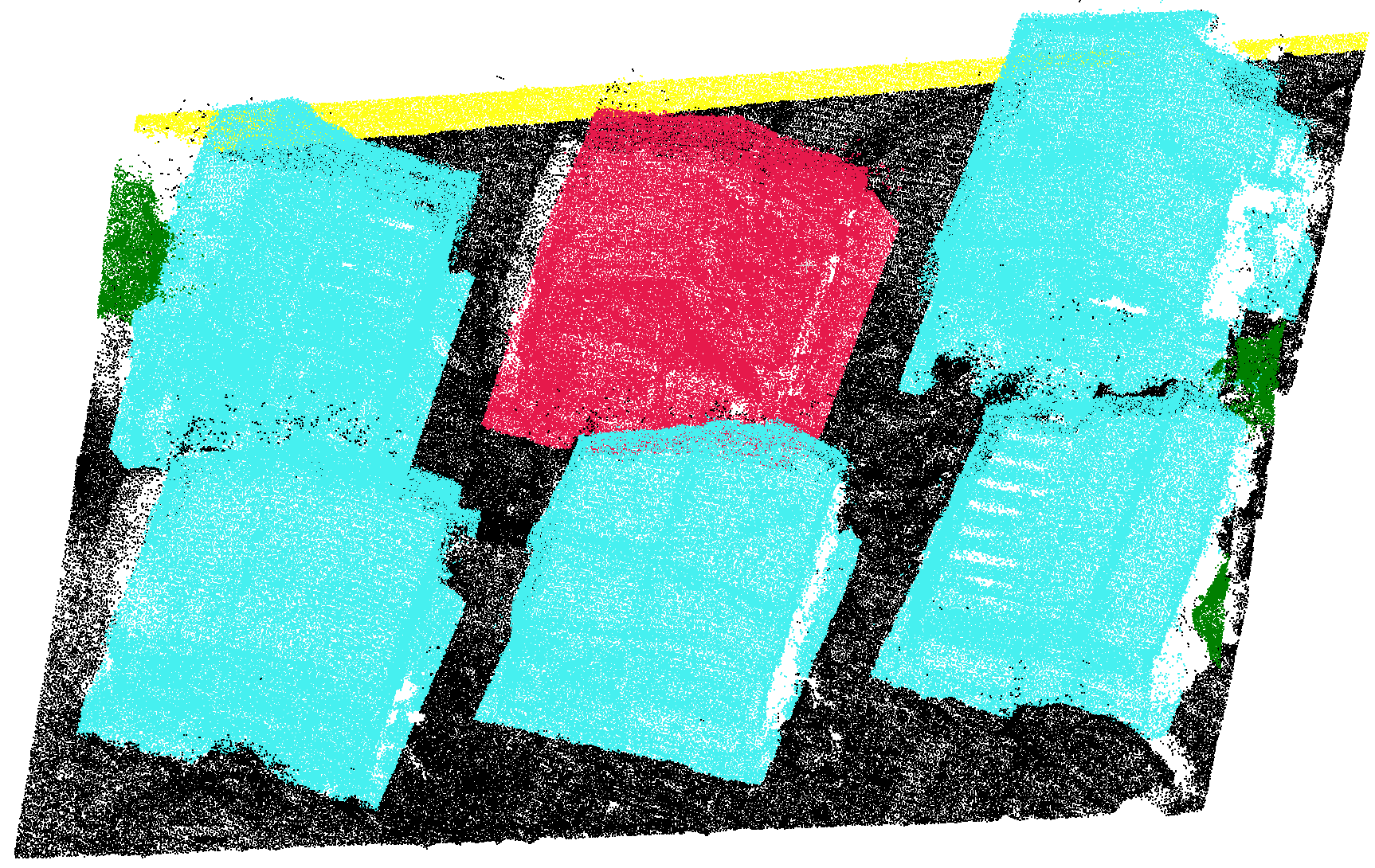}  & \includegraphics[width=0.3\linewidth]{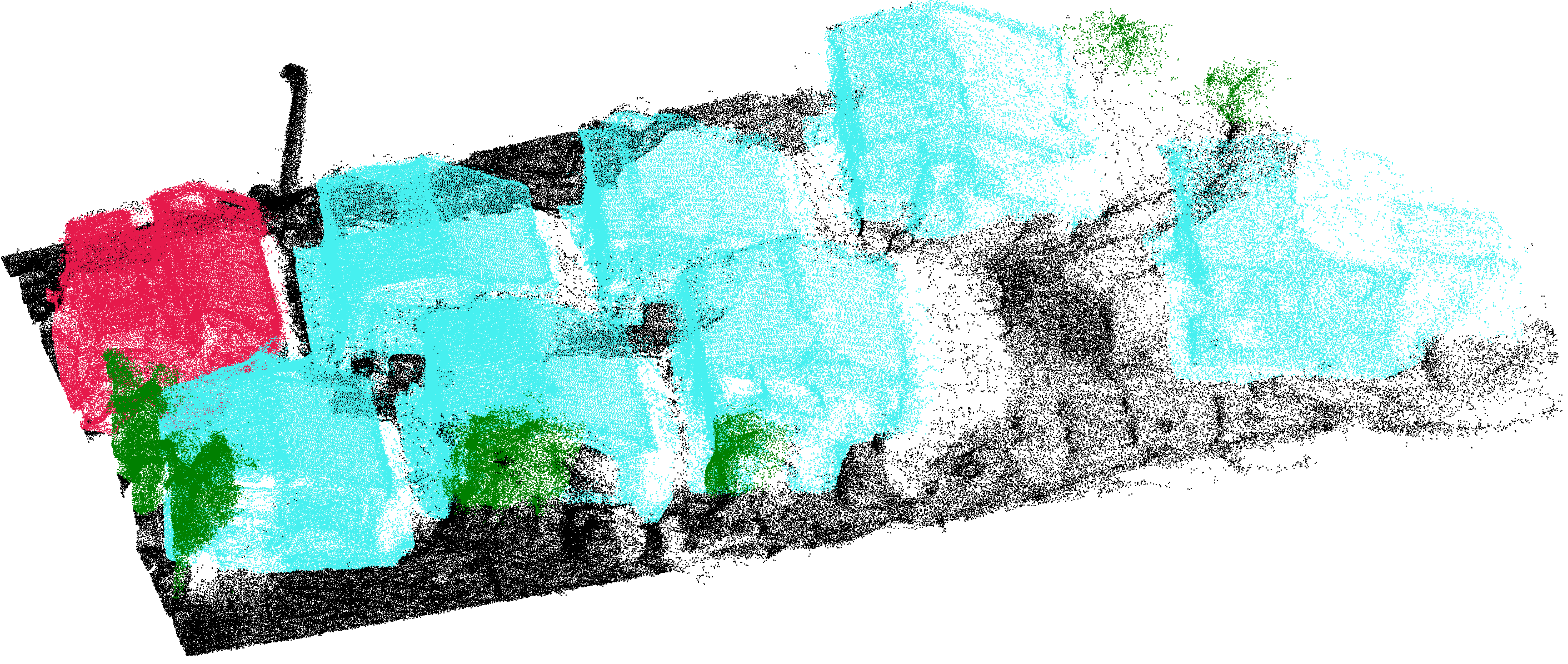} & \includegraphics[width=0.23\linewidth]{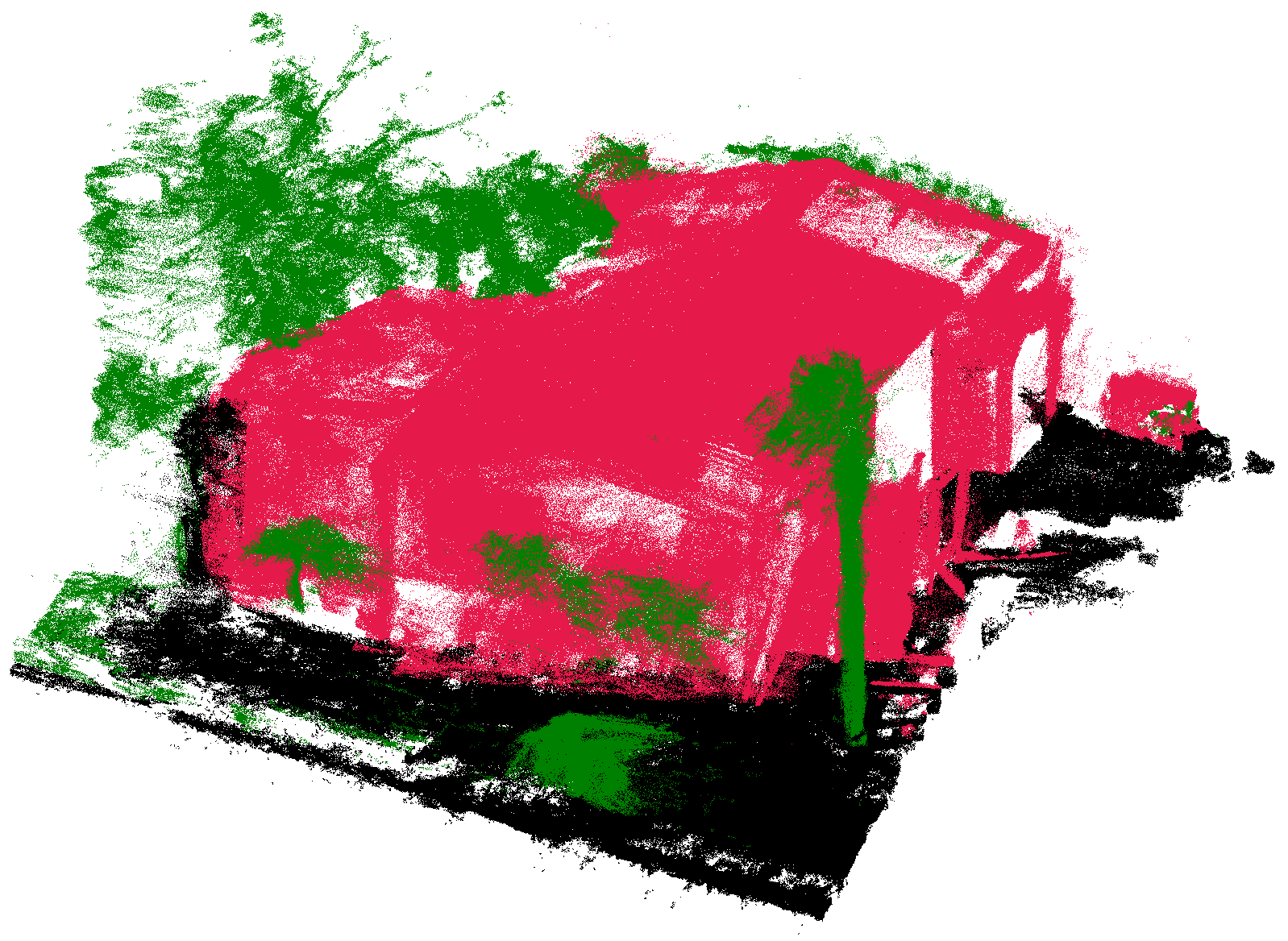}  \\
PTv1 & \includegraphics[width=0.23\linewidth]{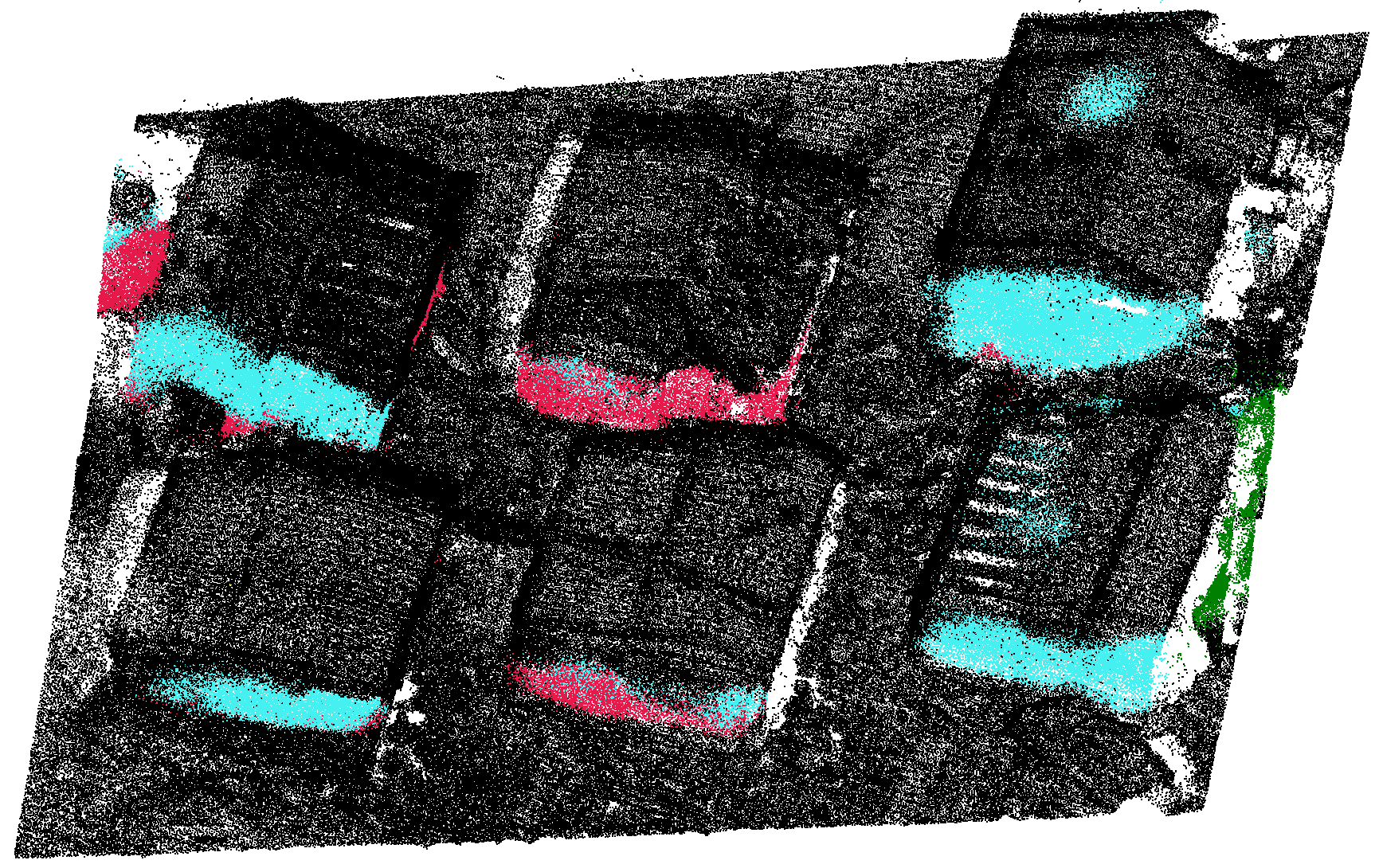}  & \includegraphics[width=0.3\linewidth]{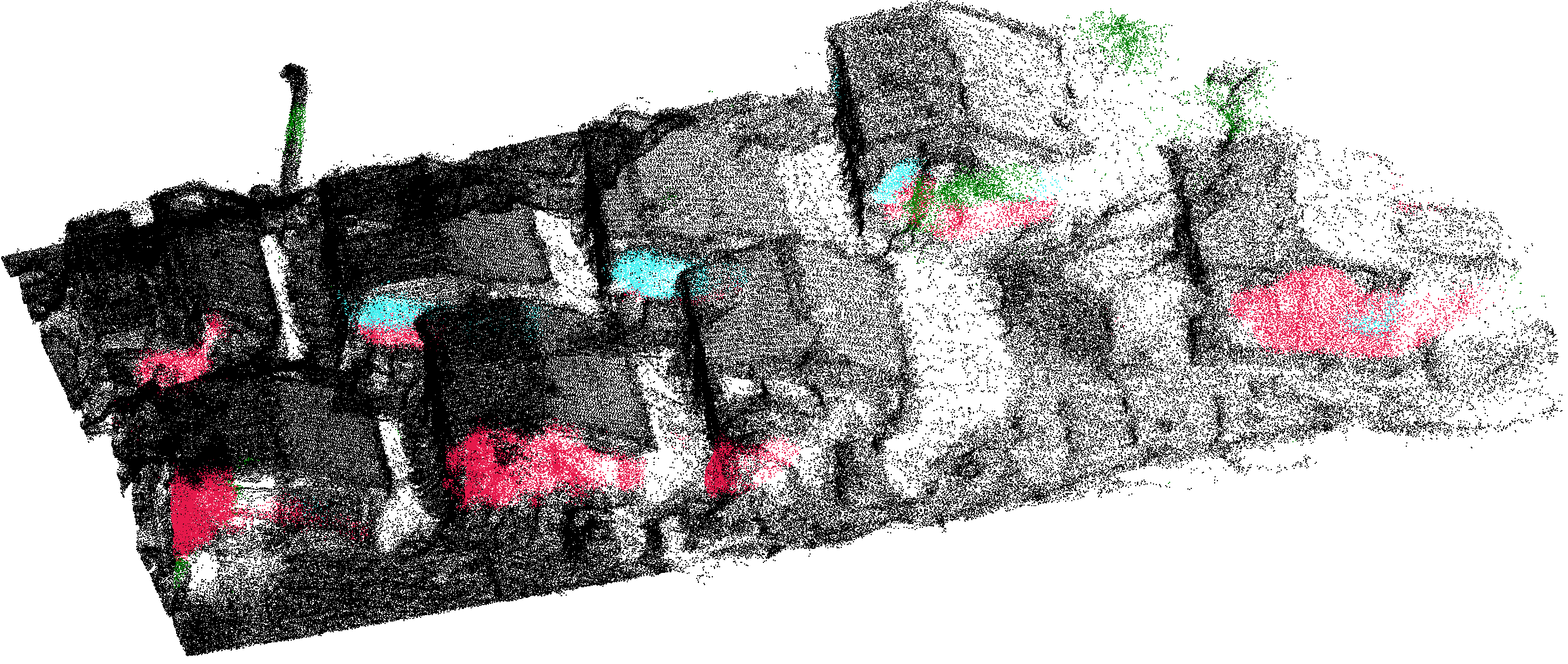} & \includegraphics[width=0.23\linewidth]{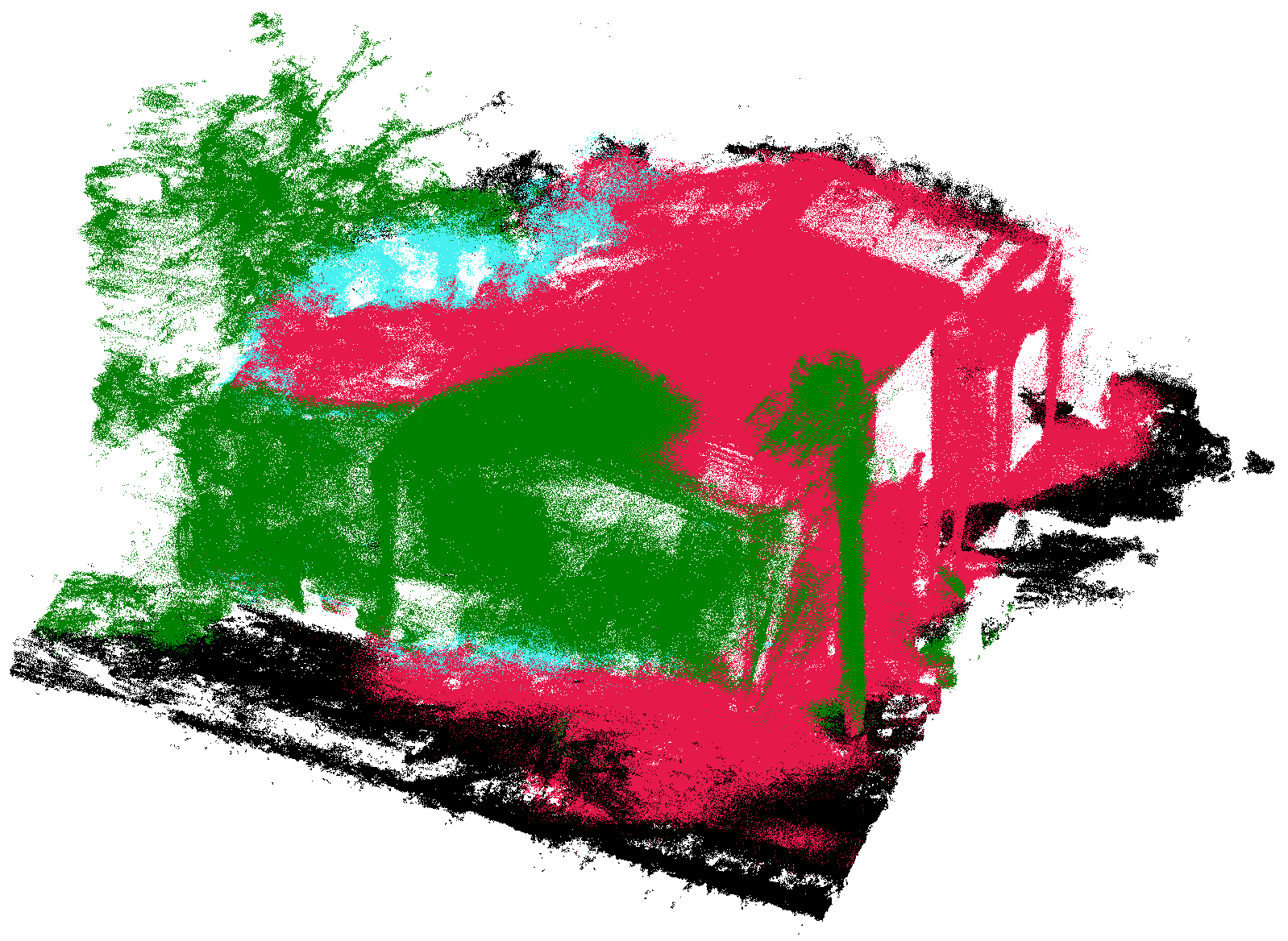}  \\ 
FPT & \includegraphics[width=0.23\linewidth]{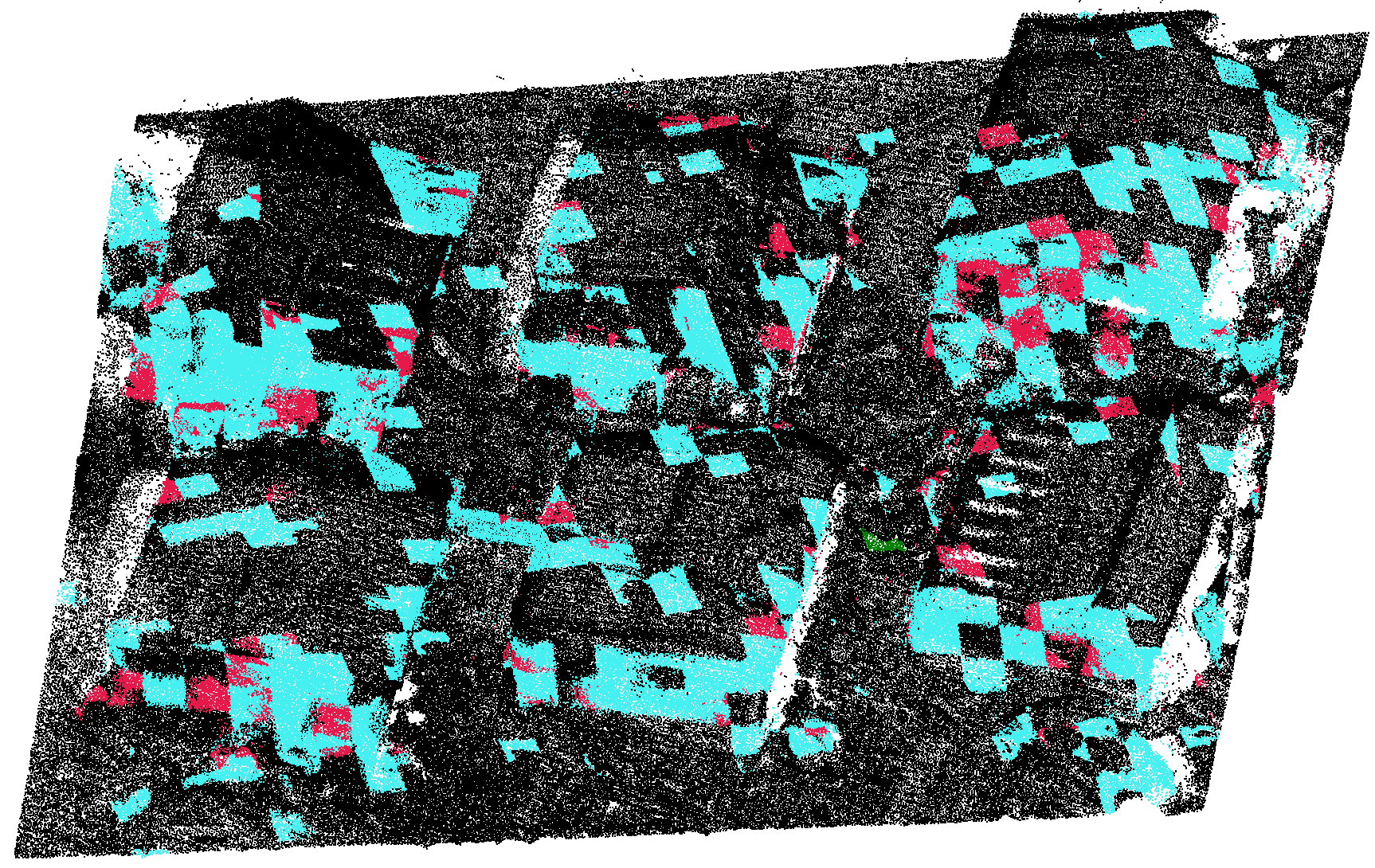}  & \includegraphics[width=0.3\linewidth]{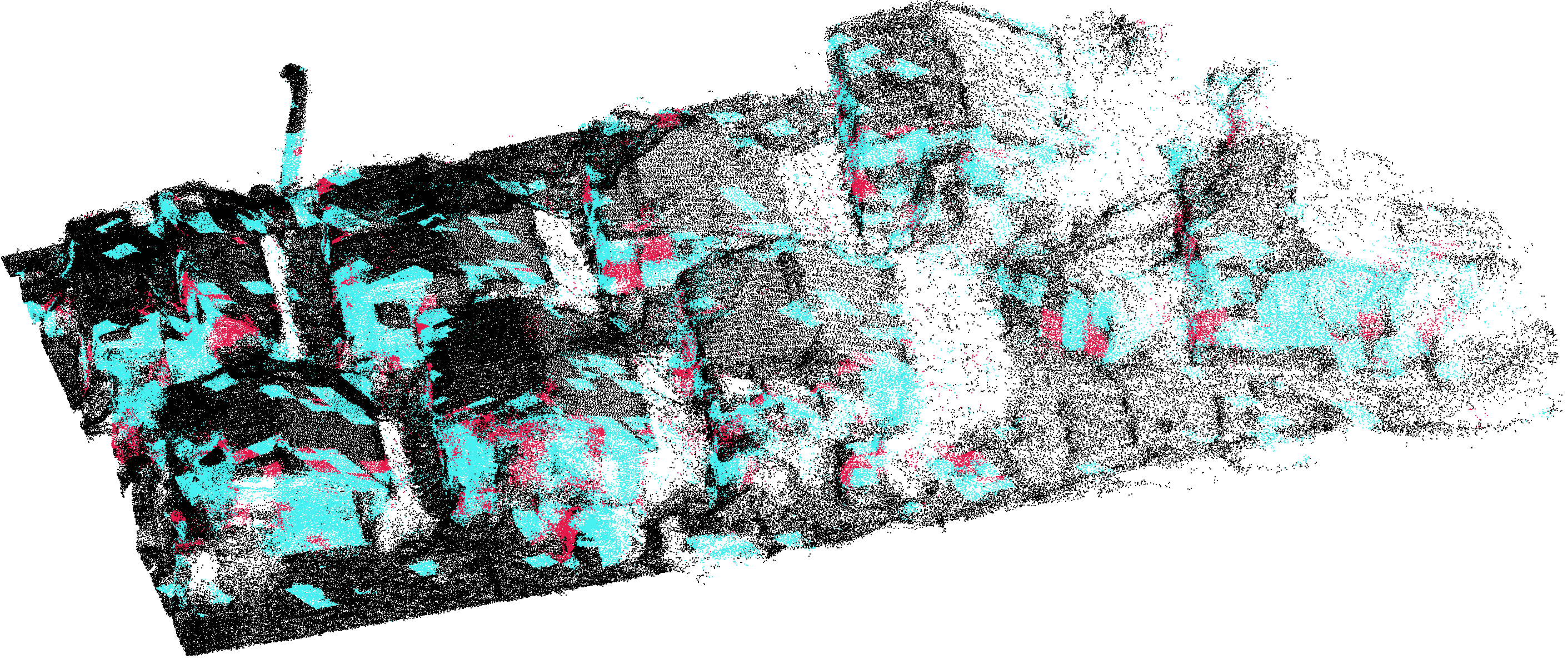} & \includegraphics[width=0.23\linewidth]{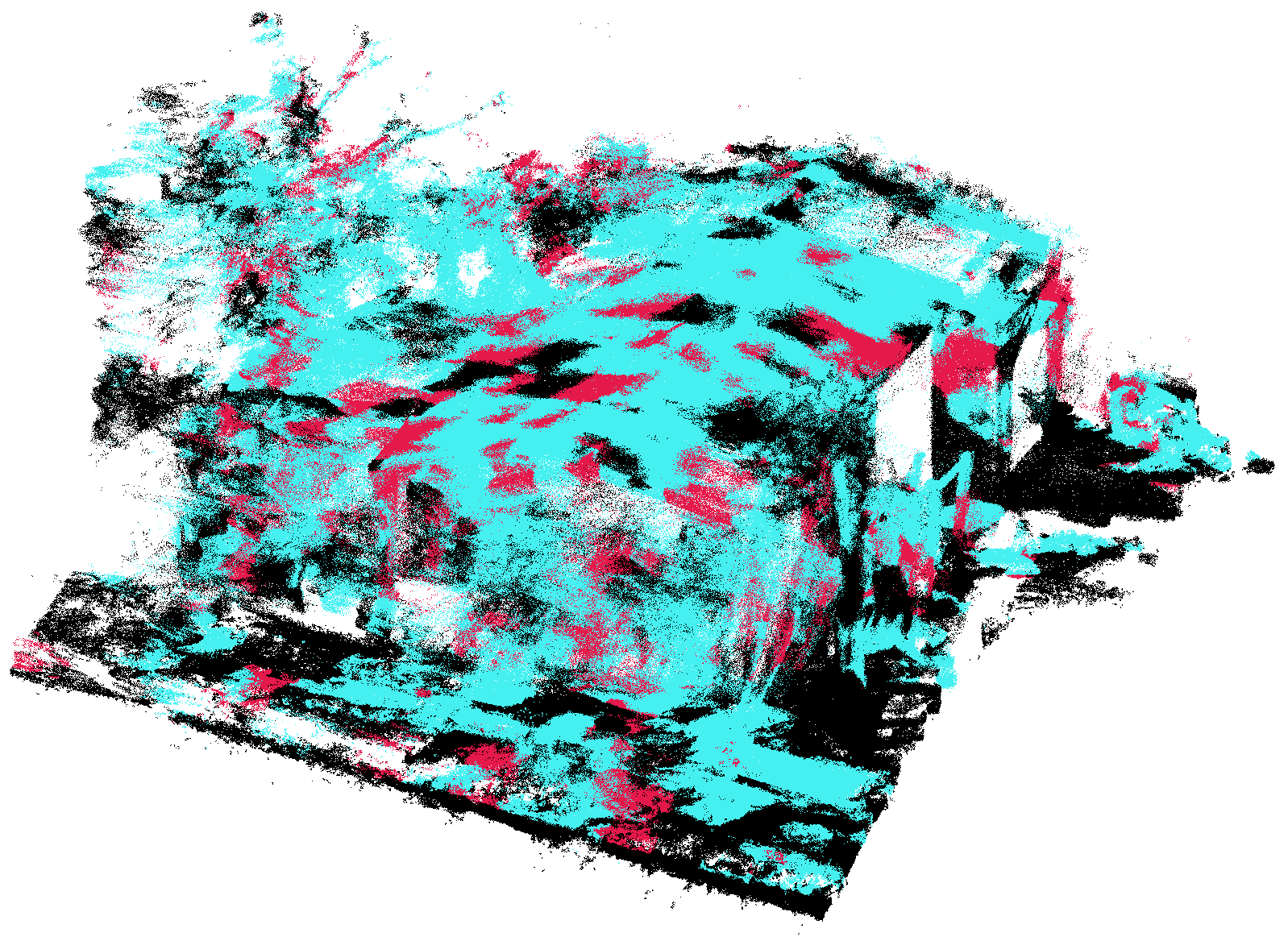}  \\ 
PTv2 & \includegraphics[width=0.23\linewidth]{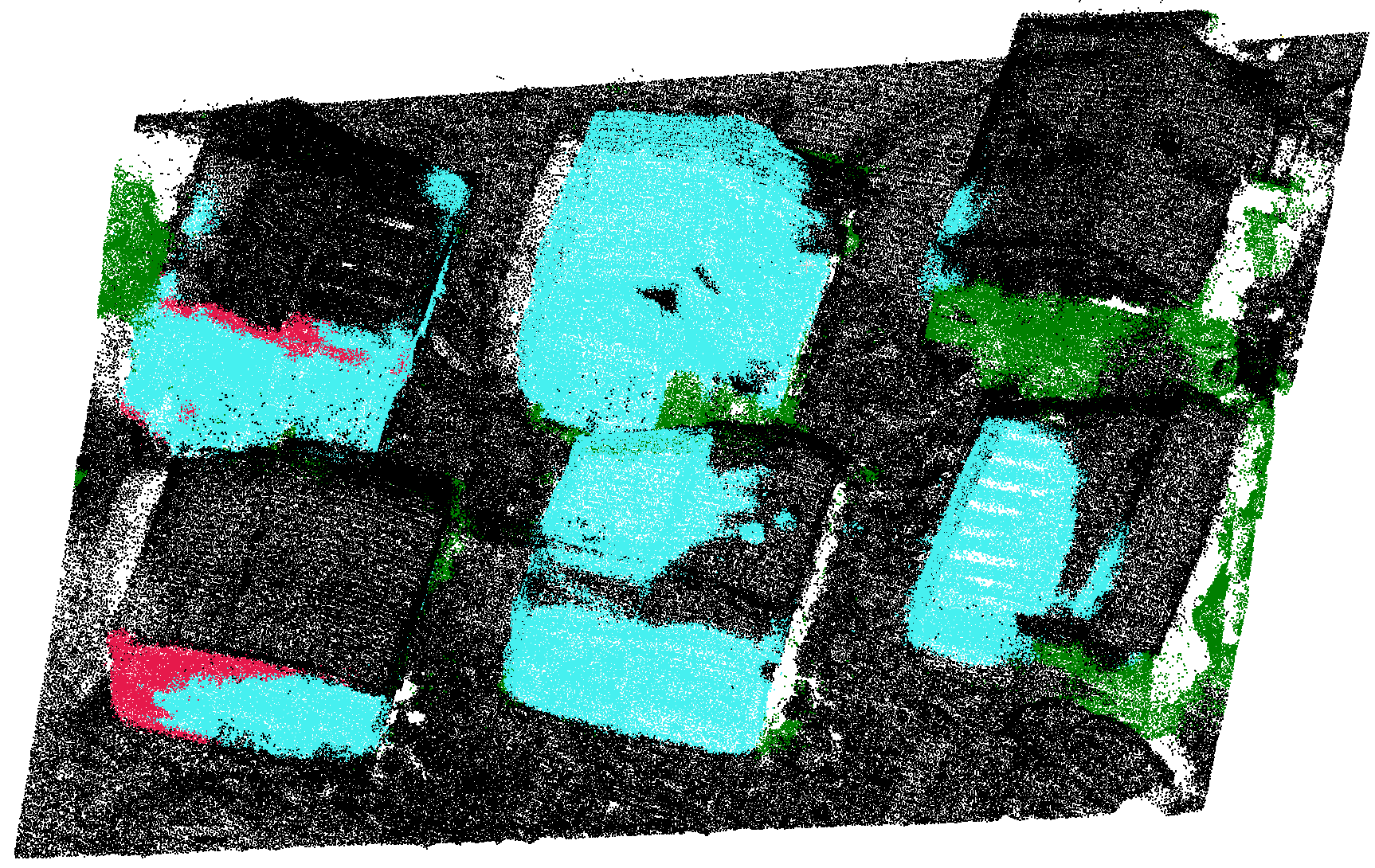}  & \includegraphics[width=0.3\linewidth]{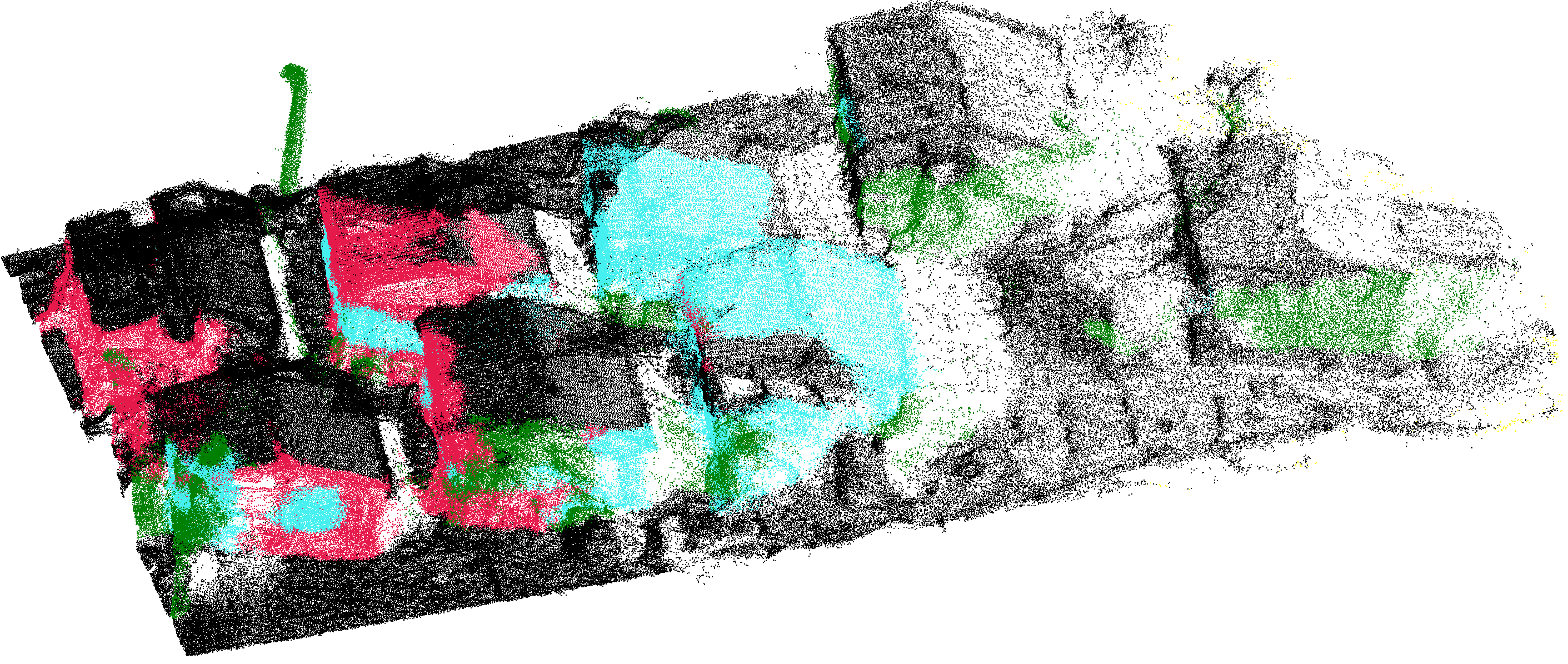} & \includegraphics[width=0.23\linewidth]{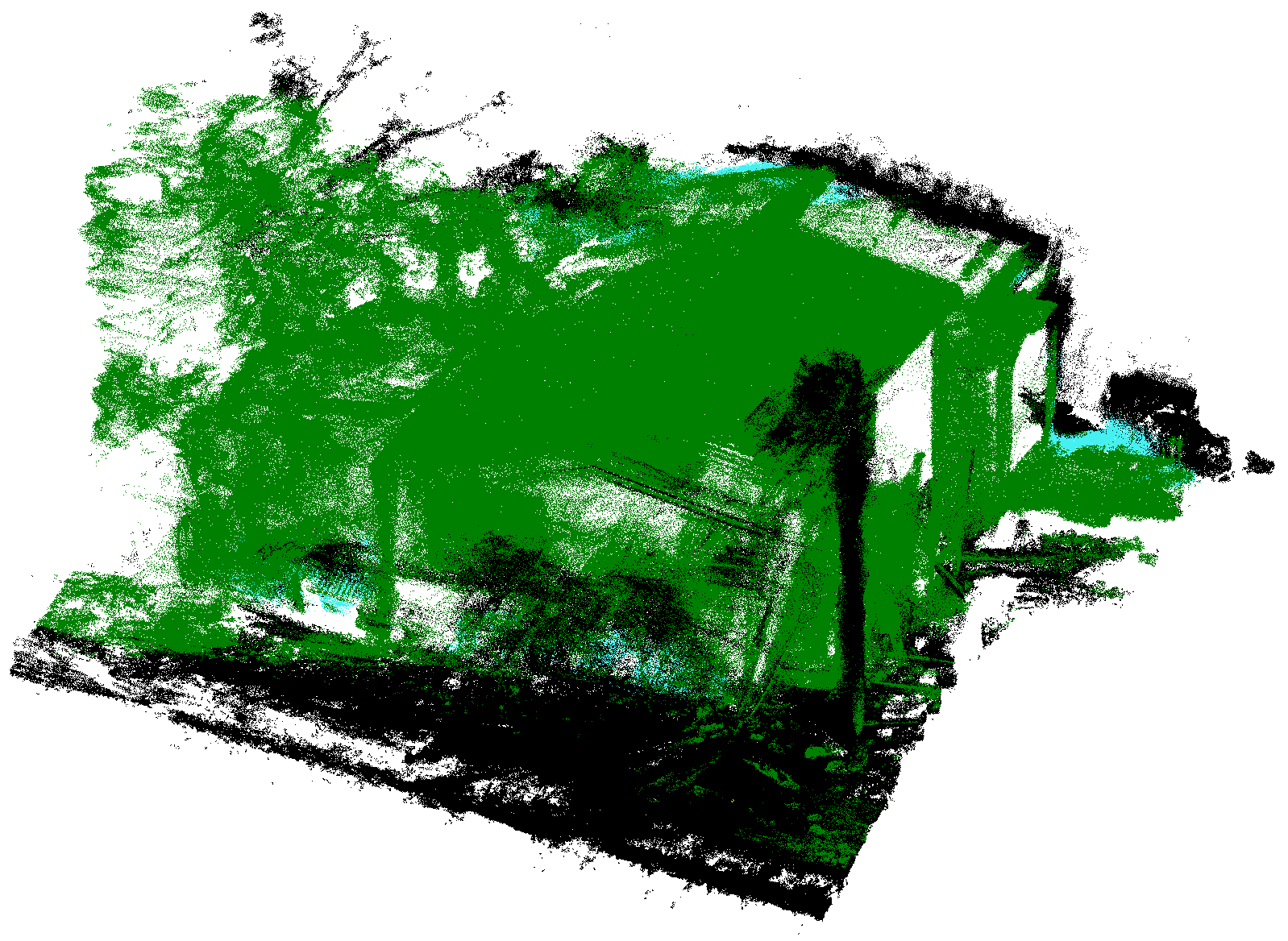}  \\ 
PTv3 & \includegraphics[width=0.23\linewidth]{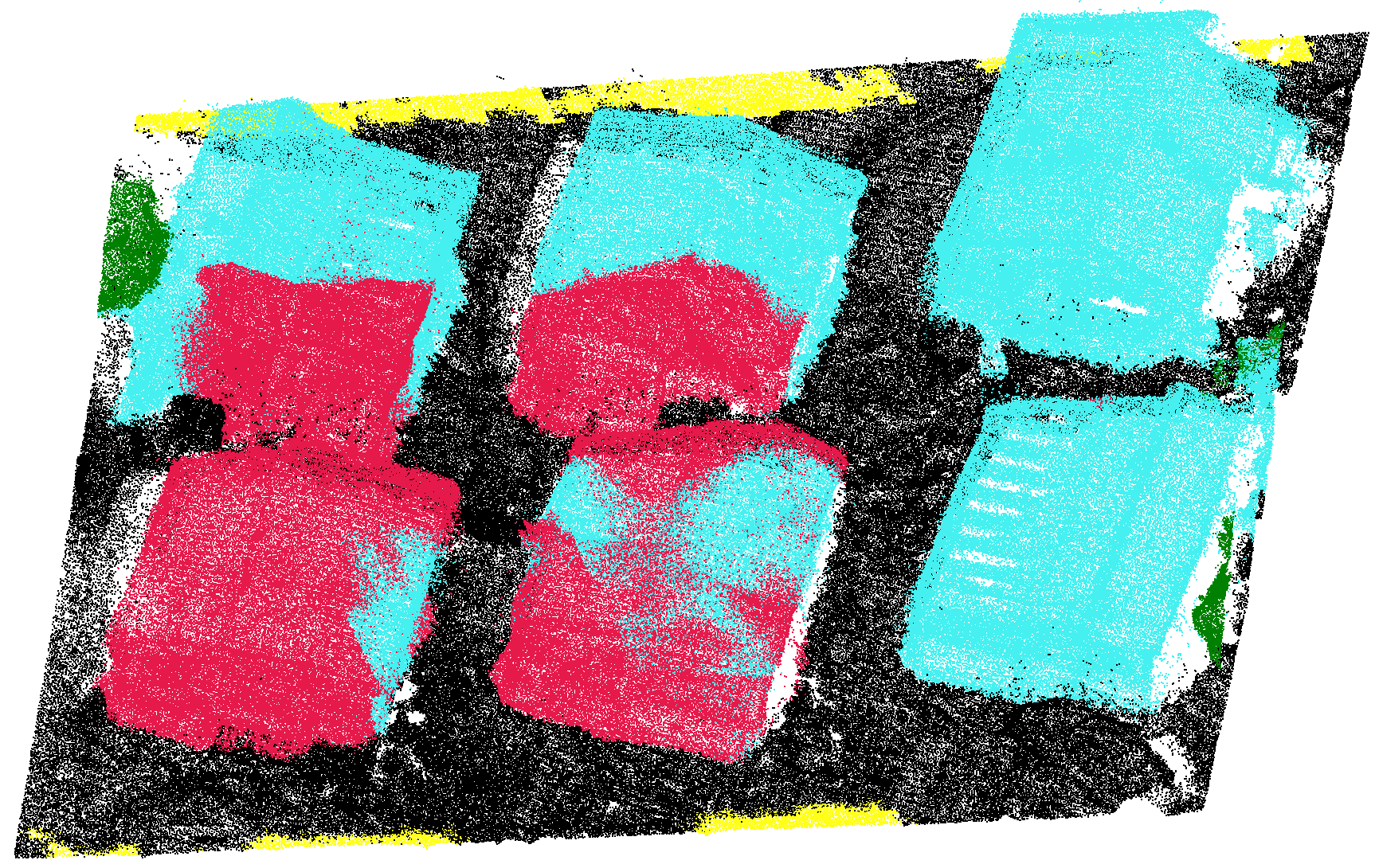}  & \includegraphics[width=0.3\linewidth]{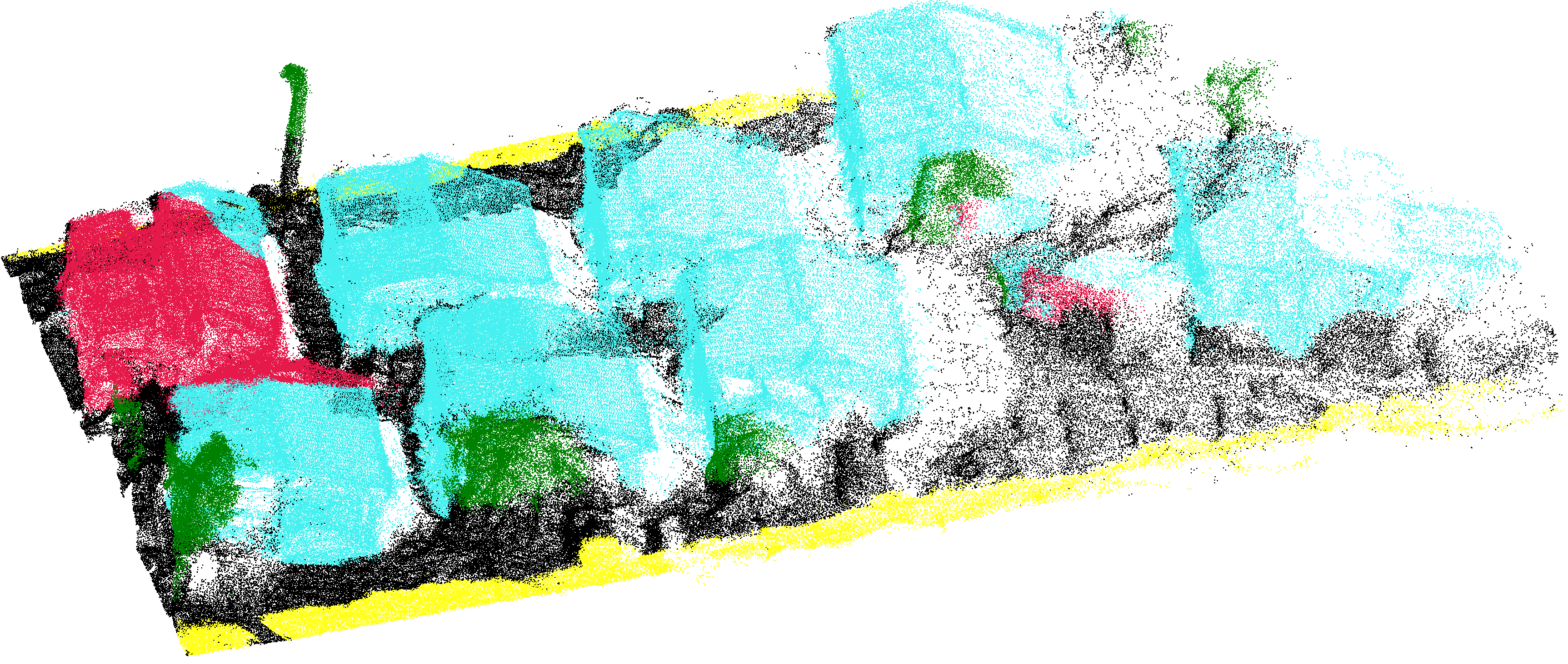} & \includegraphics[width=0.23\linewidth]{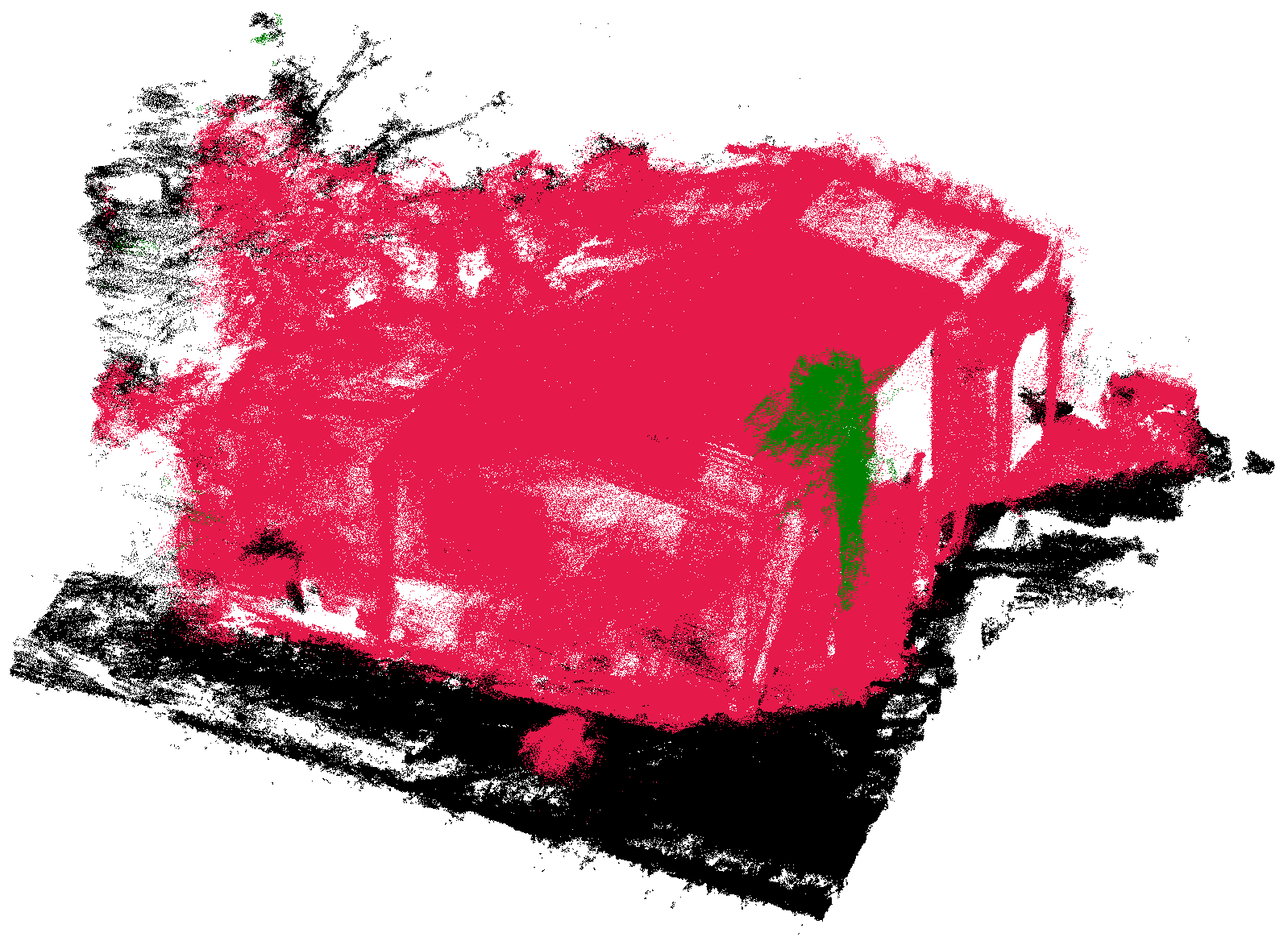}  \\ 
OA-CNNs & \includegraphics[width=0.23\linewidth]{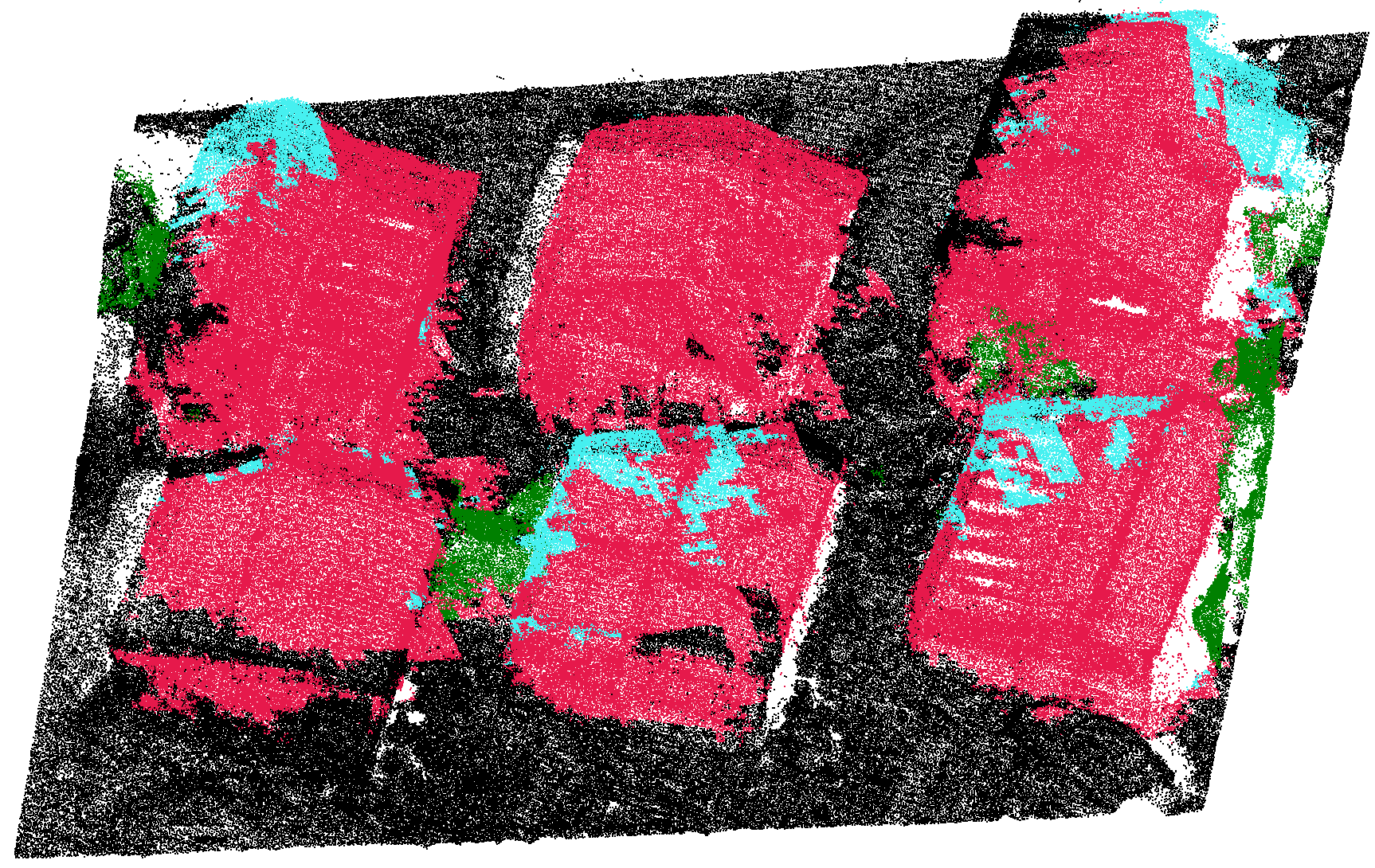}  & \includegraphics[width=0.3\linewidth]{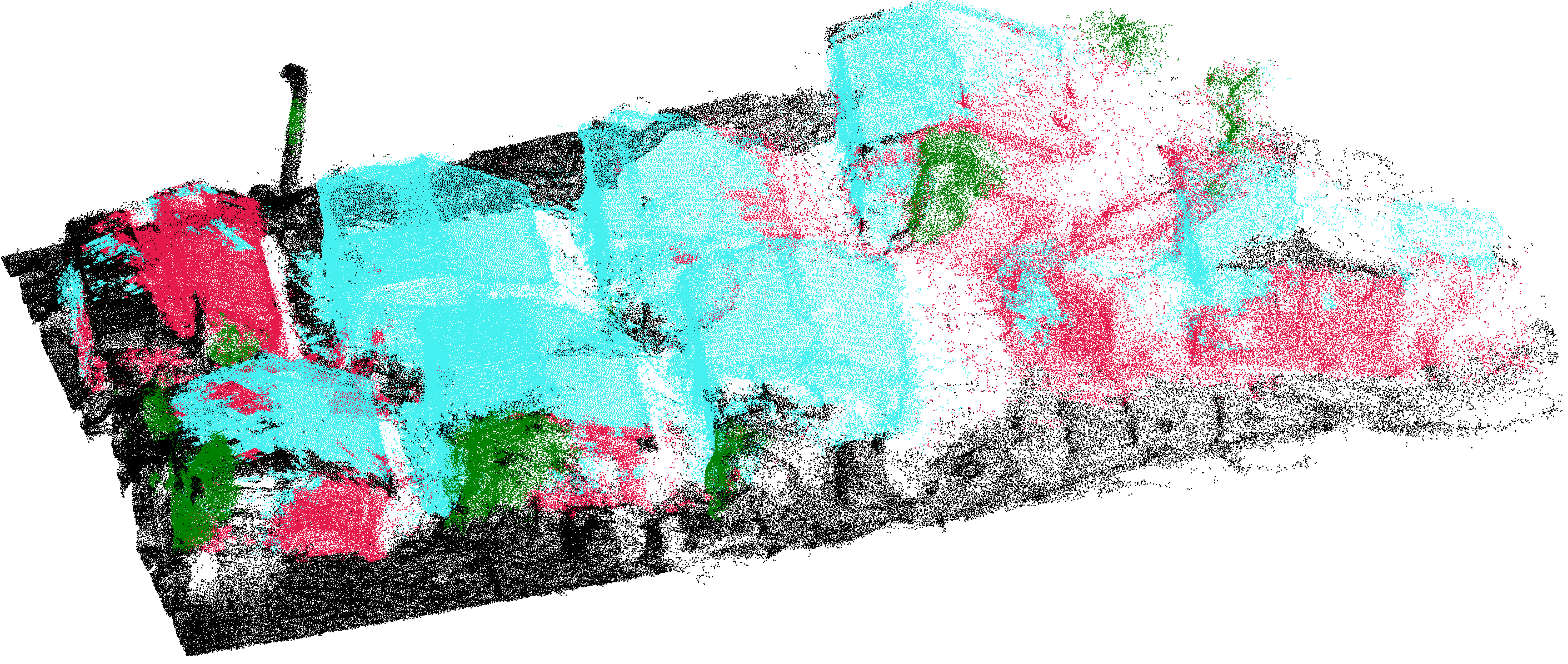} & \includegraphics[width=0.23\linewidth]{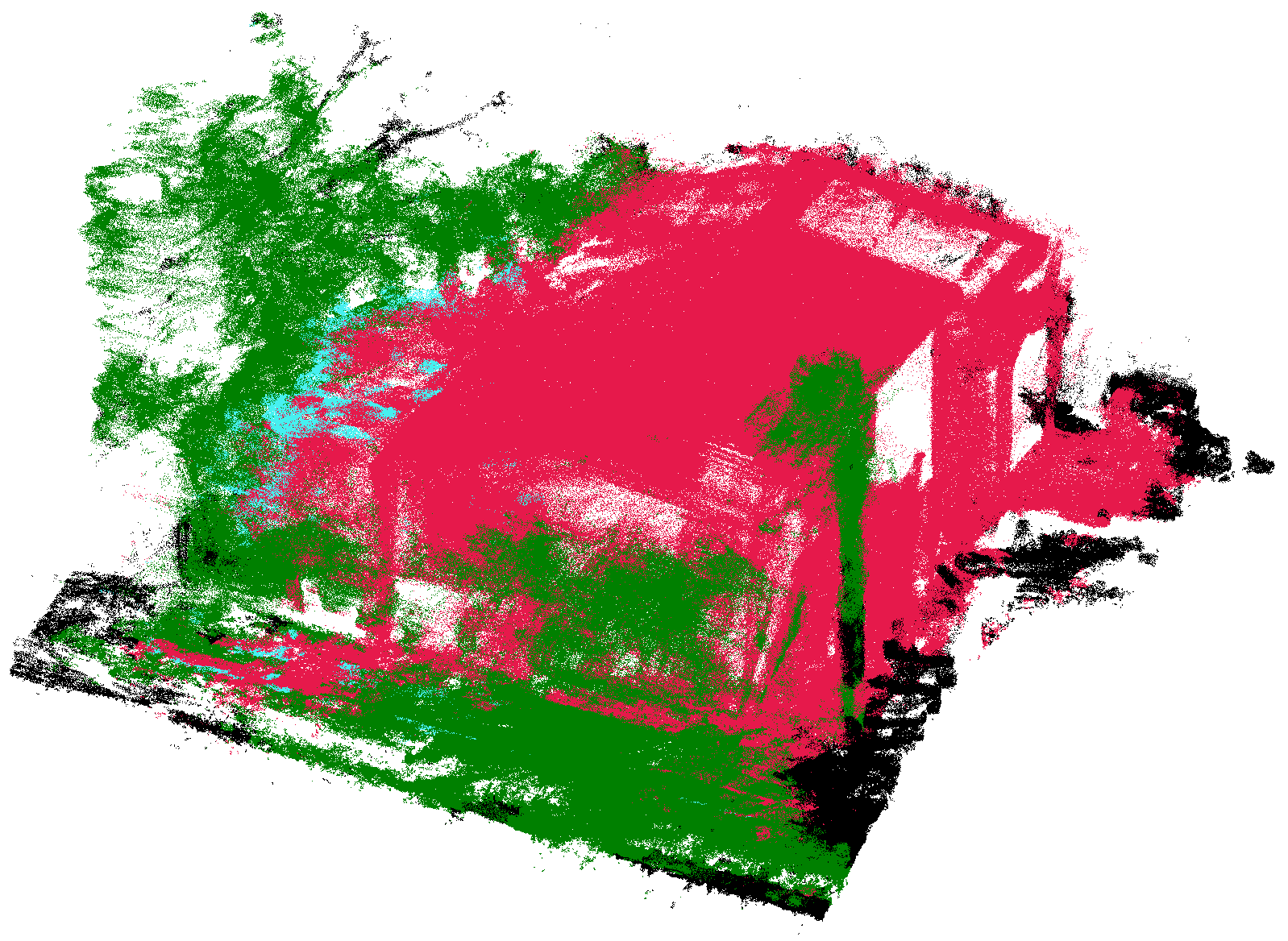}  \\ 
\end{tabular}
\label{tab:qual}
\end{table}
}

Tables \ref{tab:eval-miou} and \ref{tab:eval-macc} show the performance of SOTA 3D segmentation methods \cite{choy20194d, zhao2021point, park2022fast, wu2022point,
wu2024ptv3} on our 3D dataset. Overall, PTv3 \cite{wu2024ptv3} achieves the highest mIoU (0.4584) and mAcc (0.5508), likely due to its strong serialized modeling and multi-scale attention, which help capture both structural layout and fine-grained damage cues. For the Building-Damage class, PTv3 \cite{wu2024ptv3} leads with a large margin in mIoU (0.7614), while OA-CNNs \cite{Peng_2024_CVPR} achieves the highest accuracy (0.9276). It can show that the serialization strategy in PTv3 \cite{wu2024ptv3} and convolutional layers in OA-CNNs \cite{Peng_2024_CVPR} can detect fine-grained details more efficiently than the other baseline methods. In contrast, Building-no-Damage and Road are best handled by PTv2 \cite{wu2022point}. The improvement of positional encoding and grouped vector attentions in PTv2 \cite{wu2022point} likely contributes to its effectiveness in modeling flat, textureless surfaces. For Trees, OA-CNNs \cite{Peng_2024_CVPR} shows the highest mAcc (0.7248), while PTv3 \cite{wu2024ptv3} has the best mIoU (0.4001). PTv3 \cite{wu2024ptv3} also excels at identifying Background regions. In conclusion, PTv3 \cite{wu2024ptv3} achieves the best performance among the baseline methods due to its improved architecture in processing the input point cloud by serializing it to handle irregular shapes efficiently.
\subsection{Qualitative Result on SOTA 3D Semantic Segmentation Methods}
We present qualitative comparisons of state-of-the-art 3D segmentation methods \cite{choy20194d, zhao2021point, park2022fast, wu2022point, 
wu2024ptv3} on our test set (see Table~\ref{tab:qual}). For scenes containing a single building, most methods accurately recognize building shapes, regardless of damage condition. However, in more complex scenes with multiple buildings, several methods struggle to distinguish between damaged and undamaged structures. Specifically, models such as 
PTv3 \cite{wu2024ptv3}, and OA-CNNs \cite{Peng_2024_CVPR} often confuse these classes. Meanwhile, methods like PTv1 \cite{zhao2021point}, FPT \cite{park2022fast}, and PTv2 \cite{wu2022point} not only misclassify damage levels but also fail to reconstruct building shapes in cluttered scenes. These qualitative results highlight the difficulty of modeling fine-grained structural differences in disaster scenarios, especially when multiple buildings and occlusions are present.

\section{Code Availability}
\textbf{3DAeroRelief} can be used without any accompanying code, as no specific implementation is required.


\section*{Acknowledgements}
We gratefully acknowledge the Center for Robot-Assisted Search and Rescue (CRASAR) and the National Science Foundation (grant \#2423211) for sponsoring this work.
\section*{Authors Contributions}
\subsection*{Authors and Affiliations}
\textbf{Department of Computer Science and Engineering, Lehigh University, Bethlehem, Pennsylvania, 18015, USA}\\
Nhut Le, Ehsan Karimi, Maryam Rahnemoonfar
\\
\\
\textbf{Department of Civil and Environmental Engineering, Lehigh University, Bethlehem, Pennsylvania, 18015, USA}\\
Maryam Rahnemoonfar
\subsection*{Contributions}
Nhut Le preprocessed data, reconstructed 3D models, and partially conducted the experiments. Ehsan Karimi labeled segmentations in 2D images, refined semantic segmentations in 3D point clouds, and partially conducted the experiments. Maryam Rahnemoonfar conceived the research idea, secured funding, supervised the project, and critically reviewed and revised the manuscript.
\subsection*{Corresponding author}
Correspondence to Dr. Maryam Ranehmoonfar, email: maryam@lehigh.edu 
\section*{Competing Interests}
The authors declare no competing interests.


\end{document}